\documentclass[conference]{IEEEtran}
\usepackage{times}

\usepackage[numbers,sort]{natbib}
\usepackage{multicol}
\usepackage{hyperref}
\usepackage{graphicx}
\usepackage{amsmath,amssymb,amsfonts}
\usepackage{algorithm, algorithmic}
\usepackage{subcaption}
\usepackage{multirow}
\usepackage[export]{adjustbox}
\usepackage{dblfloatfix}
\usepackage[capitalize]{cleveref}

\usepackage{xcolor}
\usepackage{wrapfig}

\definecolor{dgreen}{rgb}{0.00,0.49,0.00}
\definecolor{dblue}{rgb}{0,0.08,0.75}
\hypersetup{
    colorlinks = true,
    citecolor=dgreen,
    urlcolor=dblue,
    linkcolor=dblue
}

\renewcommand{\paragraph}[1]{\noindent{\bfseries #1.}}

\crefname{assumption}{Assumption}{Assumptions}
\crefname{equation}{}{}
\Crefname{equation}{Eq.}{Eqs.}
\crefname{figure}{Fig.}{Figs.}
\crefname{table}{Tab.}{Tabs.}
\crefname{section}{Sec.}{Sec.}
\crefname{theorem}{Thm.}{Thm.}
\crefname{proposition}{Prop.}{Props.}
\crefname{lemma}{Lemma}{Lemmas}
\crefname{corollary}{Cor.}{Cor.}
\crefname{example}{Example}{Examples}
\crefname{remark}{Remark}{Remarks}
\crefname{algorithm}{Alg.}{Algorightms}

\crefname{appendix}{App.}{App.}
\crefname{subappendix}{Appendix}{Appendices}
\crefname{subsubappendix}{Appendix}{Appendices}

\DeclareMathOperator*{\argmin}{arg\,min}

\newcommand{\Alg}{\textrm{Alg}}
\newcommand{\ridge}{\phi}
\newcommand{\embd}{\psi}

\usepackage{booktabs}
\usepackage{arydshln}

\begin{document}

\title{Investigating Vision Foundational Models\\ for Tactile Representation Learning}

{
\footnotesize
\author{\authorblockN{Ben Zandonati\authorrefmark{1}}
\authorblockA{ \footnotesize University of Cambridge\\
baz23@cam.ac.uk}
\and
\authorblockN{Ruohan Wang\authorrefmark{1}}
\authorblockA{ \footnotesize Institute for Infocomm Research, A*STAR\\
john.rh.wang@gmail.com}
\and
\authorblockN{Ruihan Gao}
\authorblockA{\footnotesize Carnegie Mellon University\\
ruihang@andrew.cmu.edu}
\and
\authorblockN{Yan Wu}
\authorblockA{\footnotesize Institute for Infocomm Research, A*STAR\\
wuy@i2r.a-star.edu.sg}}
}

\maketitle

\begin{abstract}
Tactile representation learning (TRL) equips robots with the ability to leverage touch information, boosting performance in tasks such as environment perception and object manipulation. However, the heterogeneity of tactile sensors results in many sensor- and task-specific learning approaches. This limits the efficacy of existing tactile datasets, and the subsequent generalisability of any learning outcome. In this work, we investigate the applicability of vision foundational models to sensor-agnostic TRL, via a simple yet effective transformation technique to feed the heterogeneous sensor readouts into the model. Our approach recasts TRL as a computer vision (CV) problem, which permits the application of various CV techniques for tackling TRL-specific challenges. We evaluate our approach on multiple benchmark tasks, using datasets collected from four different tactile sensors. Empirically, we demonstrate significant improvements in task performance, model robustness, as well as cross-sensor and cross-task knowledge transferability with limited data requirements.
\end{abstract}

\IEEEpeerreviewmaketitle

\section{Introduction} \label{sec:introduction}


The sense of touch allows humans to feel, understand and ultimately manipulate through physical interaction. It is vital for exploration, object discrimination and fine-grained control, especially where visual perception lacks the resolution to detect surface changes, or is denied entirely. Inspired by the human sense of touch, robotic tactile learning has improved performance in tasks such as object/environment recognition~\cite{doi:10.1177/1729881417717056,luo2015tactile}, pick-and-place~\cite{8871350} and in-hand manipulation~\cite{https://doi.org/10.48550/arxiv.2207.02843}.


Tactile representation learning (TRL) leverages machine learning (ML) to make sense of the rich data generated by specialized tactile sensors. Design choices such as sampling resolution, operating conditions and cost result in different tactile sensors adopting distinct sensing mechanisms (e.g. visual signals~\cite{Yuan2017gelsight} and barometric signals~\cite{fishel2012sensing}). Ideally, TRL should be sensor-agnostic, accommodating various data formats of different sensors and able to construct consistent representations of objects and environments. In practice, however, most methods developed are sensor-specific with tailored architectures and data processing routines~\citep[e.g.][]{Yuan2017gelsight,khamis2018papillarray,9340693,9448230}.

This siloed approach has multiple limitations. First, individual tactile datasets are usually small due to the high cost of data collection. The tactile representation derived from such small datasets often generalize less well, especially for out-of-distribution data~\citep[e.g.,][]{9341111,9340693}. Even calibration differences and expected wear from regular usage present domain shifts detrimental to model performance. Furthermore, the lack of a unifying data format for different tactile sensors makes it difficult to reuse knowledge captured in learned representations. For a new sensor design, the accompanying tactile representation model has to be learned from scratch, along with expensive data collection. All these limit the effectiveness and efficiency of TRL.

The above limitations are further highlighted when we contrast TRL with other application domains like computer vision (CV), and natural language processing (NLP). Both CV and NLP benefit from a unifying input format (images and text respectively), which permits fully shared model architectures for convenient knowledge transfer. In particular, foundational models~\cite{bommasani2021opportunities} are trained on massive datasets such as ImageNet~\cite{imagenet_cvpr09} and CommonCrawl~\cite{raffel2020exploring} to derive general representational knowledge, which can be specialized to diverse downstream tasks, such as semantic segmentation~\cite{bardes2022vicregl} in CV, and sentiment analysis~\cite{brown2020language} in NLP. Foundational models improves learning efficiency and model robustness of downstream tasks, especially for limited training data~\cite{brown2020language}.


Biologically, the human somatosensory system shares similar neural mechanisms with the visual cortex responsible for processing spatial features~\cite{doi:https://doi.org/10.1002/9781118133880.hop203008}. This implies that tactile properties such as texture are largely descriptions of surface spatial properties~\cite{Haindl2013VisualTA}, motivating the question of whether \textit{a vision foundational model could be exploited to tackle the aforementioned challenges in TRL.}
Specifically, we investigate the following:
\begin{itemize}
    \item Can vision models be agnostic to data from heterogeneous tactile sensors?
    \item Can vision foundational models improve model performance and robustness for TRL?
    \item Can vision architecture facilitate efficient knowledge transfer between downstream learning tasks and models trained on different sensor data?
\end{itemize}


In this work, we present a unified approach to address the above questions. We first present the use of \textit{tactile images} as a simple unifying data format for heterogeneous tactile sensory outputs, to encode them as spatial features. This recasts TRL as a vision task, but with different input image sizes for different sensors. We adopt convolutional models~\cite{krizhevsky2017imagenet} as the fully shared architecture for all sensors, exploiting convolution's agnosticity to image sizes.

The above construct enables efficient knowledge transfer in multiple ways. First, we show that a foundational vision model pre-trained on natural images can be directly applied to tactile learning tasks by simply performing least square regression to the last layer, providing evidence on the connection between visual and tactile perception in a non-biological system. Second, the foundational model can also be fine-tuned into tactile representation models with improved performance and robustness. In particular, we leverage data augmentation to counteract the limited tactile data during fine-tuning. Lastly, we demonstrate that the fine-tuned tactile representation model retains general features to allow cross-task and cross-sensor transfer.

To evaluate our proposed approach, we consider multiple benchmark tasks including standard material classification, continual learning for material classification and detection of fabric composition. We specifically test on data collected from four different sensors, with different data collection procedures, to demonstrate the general applicability of our approach.

\paragraph{Contributions} Our key contributions are summarized below:
\begin{itemize}
    \item We extensively investigate on the feasibility, effectiveness, efficiency and robustness of using a vision foundational model for TRL. We use tactile images as a unified model input transformed from any tactile sensors.
    \item We introduce a new evaluation benchmarks for tactile learning, namely fabric composition detection.
    \item We contribute two new tactile datasets, including a material classification dataset using GelSight sensor and a fabric composition dataset using Contactile sensor.
    \item Empirically, we demonstrate that our proposed approach learns robust models for all sensors evaluated and outperforms baseline models tailored to specific sensors.
\end{itemize}

\section{Preliminaries and Related Work} \label{sec:related_work}
We present three task settings to support the comprehensive evaluation of our proposed approach. The first two tasks are standard benchmarks for TRL while the third one is a novel task of composition detection task. We also review relevant works.


\subsection{Tactile Representation Learning Tasks}
\label{sec:bg_tasks}
\paragraph{Material Classification} This is a common benchmark for TRL~\citep[e.g.][]{fishel2012bayesian, hoelscher2015evaluation, taunyazov20event, baishya2016robust, 9636380, 9340693, 5756488}. Similar to image classification, material classification determines the source material measured by a tactile sensor, from a finite number of classes. For example, early research involved classification of the textural information gathered via sliding an electret microphone across the surface of materials~\cite{727512}. The task remains a standard benchmark amid the rapid development of different sensor designs.

A natural extension to standard material classification investigates the learned model's robustness to out-of-distribution data. This includes varying data length and the moving speed of the tactile sensor (as controlled by a robot). For example, \cite{vibrotactileafferents} achieved improved robustness to the sensor's movement speed via additional sensing modalities. \cite{9340693} also proposed a customized spiking neural network to reduce the data length needed for classification.

\paragraph{Continual Learning for New Materials}
For real-world applications, robots are expected to continuously learn and adapt to novel environments. This also applies to TRL and was investigated in \cite{soh2012online, soh2014incrementally}, where robots learn new objects continuously by touch. In this work, we similarly extend material classification to the continual learning (CL)~\cite{de2021continualsurvey} setting. Formally, let $\mathrm{D} = \{B_1, B_2, \dots, B_T\}$ be a data sequence with $B_t$ denoting the data for material $t$. We wish to design a CL algorithm $\Alg(\cdot)$ in the form of
\begin{equation}
\label{eq:cl}
    (f_{t},~M_t) = \Alg(B_t, f_{t-1}, M_{t-1}),
\end{equation}
where $f_t$ is the current classification model after learning the novel material $t$. $f_t$ should be capable of classifying all materials observed so far (i.e., $B_1$ through $B_t$). A small memory buffer $M$ is allowed to store data about previous materials to mitigate model forgetting. $M_t$ denotes the current content of the memory buffer.

Intuitively, the CL algorithm $\Alg(\cdot)$ must learn each material sequentially. It also cannot access training data for previous materials except for those stored in the memory buffer. The algorithm is thus forced to learn new materials on the fly without forgetting its existing knowledge. In contrast, standard material classification learns all materials in $\mathrm{D}$ concurrently and with unlimited access to all data. CL thus represents a more challenging and realistic benchmark. 


\paragraph{Fabric Composition Detection} We introduce a new evaluation benchmark for TRL. Concretely, we design a fine-grained fabric composition detection task, in which the learned tactile model must predict the constituents of a specific fabric material, instead of simply identifying it. This task serves as a more challenging benchmark compared to standard material classification. It also allows us to investigate knowledge transfer between sensors and tasks  (e.g., from material classification to constituents detection). We will describe the new dataset collected for this task in \cref{sec:sensor_data}.

\subsection{Existing Methods} There exists a wide range of tactile sensor designs leveraging various sensing modalities, including strain gauges~\cite{5756488}, piezo-resistive layers~\cite{stag}, accelerometers~\cite{5752872}, capacitive~\cite{schmitz2010tactile}, optical~\cite{WardCherrier2018TheTF, Yuan2017gelsight} and those combining multiple sensing mechanisms~\cite{6631001, Nicholas2008BioTac}. Most tactile learning methods tailor their respective model architectures and learning algorithms to the specific sensors used~\citep[e.g.,][]{9340693,9636380,8793967,5756488}. These existing approaches learn sensor-specific mappings from raw sensor output to some latent representation, and adjust the model size based on size of sensor output. These tailored decisions inevitably lead to a siloed state for TRL: the developed models can't be easily reused for different sensors, even when the desired ML task remains identical.

\cite{9341111} partially addresses the above issues by learning a shared latent representation for two different sensors. This approach demonstrates improved performance compared to independently learning each sensor's data. However, it must still learn sensor-specific mappings from raw data to the shared representation, thus limiting its reuse potential for additional sensors. In contrast, our proposed approach standardises the transformation to map any raw sensor data to tactile images, to be processed by a fully shared ML model. As we will demonstrate in our experiments, our approach grants more flexibility towards knowledge transfer.


\section{Sensors and Datasets}
\label{sec:sensor_data}
We present the sensors and the associated datasets considered in this work. They are intended to validate the general applicability of our approach, and to contextualize the challenge posed by heterogeneous sensors. Each dataset is used for one or more learning tasks described in \cref{sec:bg_tasks}.

\paragraph{RoboSkin} Roboskin is a capactive sensor designed for iCub~\cite{schmitz2010tactile}. 
\citet{8793967} collected a material classification dataset using the RoboSkin on the iCub robot forearm, sweeping across multiple materials without strict control of velocity and exerted forces. This public dataset contains 20 different materials with 50 samples in each class. Each sample contains 75 sensor readings.

\paragraph{BioTac} SynTouch BioTac\textsuperscript{\textregistered{}} is a multi-modal tactile sensor using fluid pressure sensor and thermistor~\cite{kappassov2015tactile}. \citet{9341111} released a material classification dataset using the BioTac sensor fitted as an extended end-effector on a KUKA LBR iiwa 14 robot arm, sliding laterally across different materials with controlled speed and contact force. BioTac-20 dataset contains the same 20 materials as the RoboSkin dataset with 50 samples in each class. Each sample contains 400 readings. A larger BioTac-50 dataset was later released.


We contribute two new datasets using alternative sensors. We will release both datasets publicly to support future research in the community.

\begin{figure}[h]
    \centering
    \begin{subfigure}[b]{0.39\linewidth}
        \begin{subfigure}[b]{1\linewidth}
          \centering
          \includegraphics[width=\linewidth]{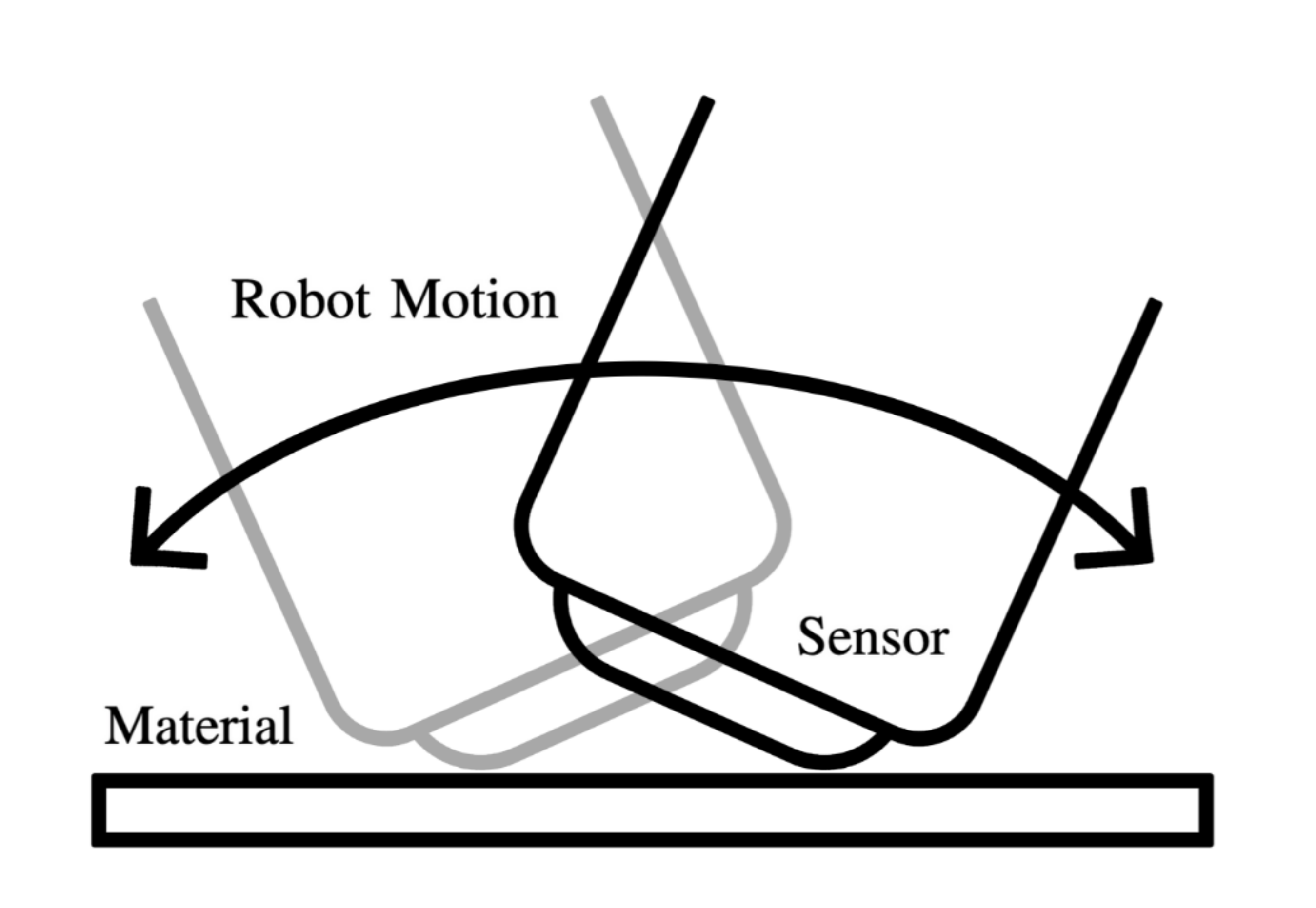}
          \caption{GelSight}
          \label{fig:gelsight_collection}
        \end{subfigure}
        \vfill
        \begin{subfigure}[b]{1\linewidth}
          \centering
          \includegraphics[width=\linewidth]{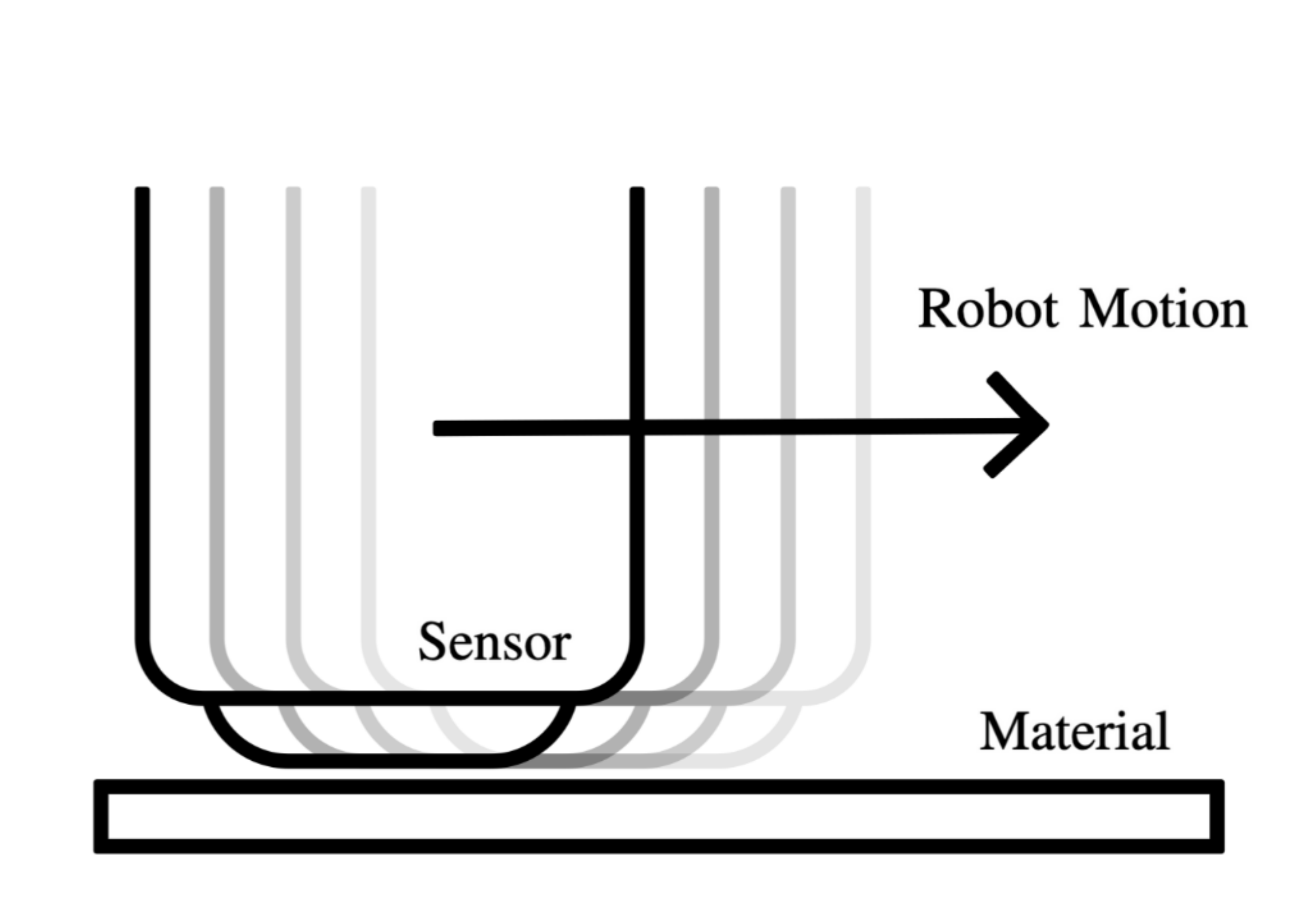}
          \caption{BioTac, RoboSkin, \\ Contactile}
          \label{fig:others_collection}
        \end{subfigure}
    \end{subfigure}
    \hfill
    \begin{subfigure}[b]{0.59\linewidth}
        \centering
        \includegraphics[width=1\linewidth]{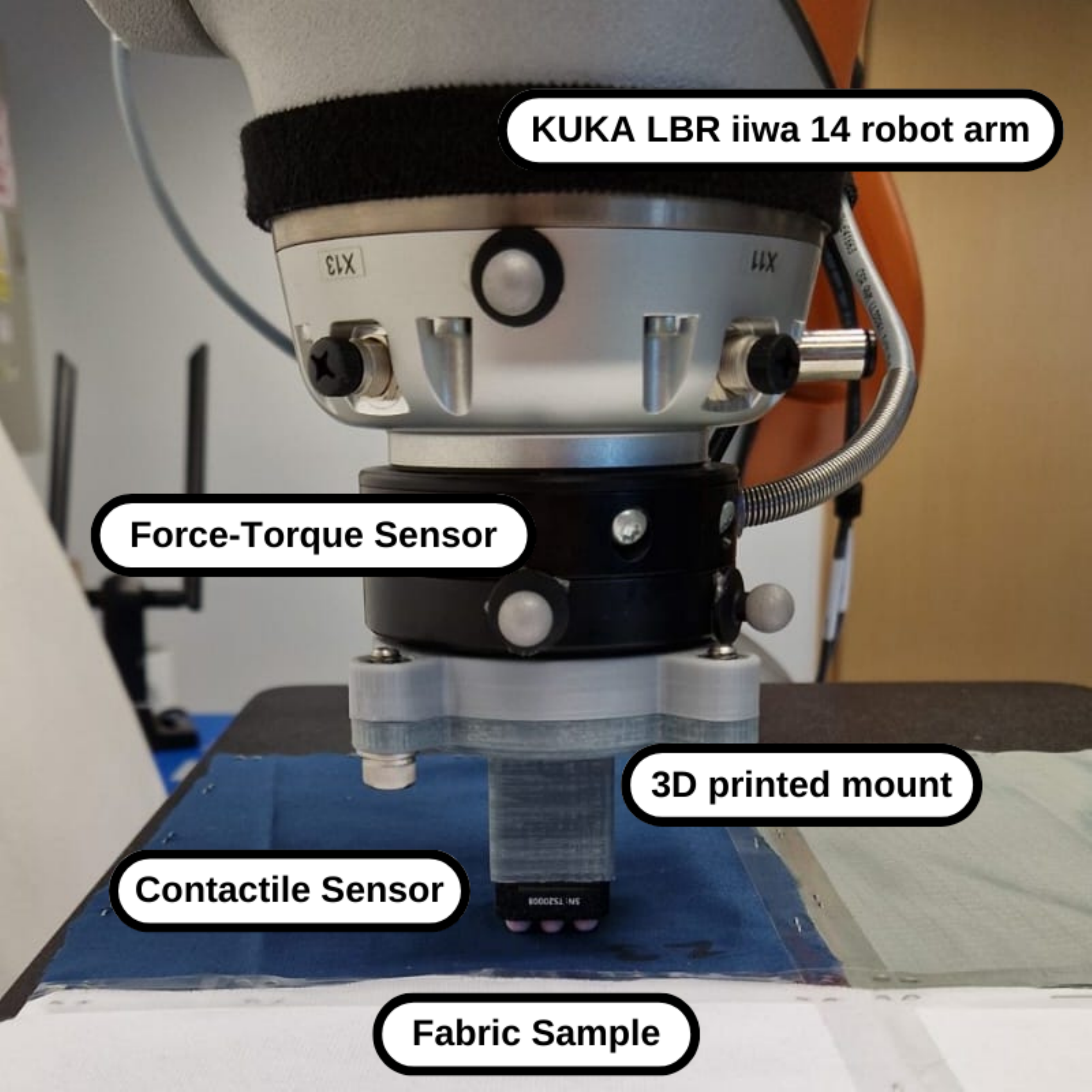}
        \caption{Setup for Contactile Protocol 1}
        \label{fig:contactile_setup}
    \end{subfigure}
    \caption{(a) and (b) are illustrations of tactile data collection process for different sensors. (c) is the robot setup for Protocol 1 in Contactile data collection. }
    \label{fig:collection_strats}
\end{figure}

\paragraph{GelSight} Gelsight is a camera-based sensor producing images of the contact surface, showing surface geometry and deformation with a soft elastomer~\cite{Yuan2017gelsight}. Each reading is an image of $480 \times 640$. A material classification dataset consists of 45 materials with 50 samples in each class. As the elastomer is vulnerable to abrasion from sliding motion, data is collected by rolling the sensor locally on material surfaces. The sensor, mounted on a KUKA LBR iiwa 14 robot arm, touches the material surface from above with a 1N force threshold. The sensor is then rotated clockwise by 1 degree, anticlockwise by 2 degrees, and finally clockwise by 1 degree back to the centre position (illustrated in \cref{fig:gelsight_collection}).


\paragraph{Contactile} Contactile\textsuperscript{\textregistered{}} sensor uses a soft, silicone array based on PapillArray~\cite{khamis2018papillarray}. The sensor measures deflection, force and vibration. We collect the data using two protocols. Protocol 1 is identical to that of BioTac dataset. In Protocol 2, the sensor is handheld and slid across materials casually with different contact forces, speeds and along different directions, to mimic more realistic and natural movements. The dataset contains samples collected from 32 fabrics, each consisting of possible 6 constituent materials: Linen, Viscose, Cotton, Wool, Polyester and Elastane (see \cref{tab:fabric_samples} for examples). 40 and 10 samples per material are collected for Protocols 1 and 2 respectively. The collection setup is illustrated in \cref{fig:others_collection,fig:contactile_setup}.




\begin{table}[h]
\centering
\caption{Fabric examples and their composition materials}
\label{tab:fabric_samples}
\resizebox{\columnwidth}{!}{%
\begin{tabular}{lllllll}
\midrule
\multicolumn{1}{l}{\multirow{2}{*}{\textbf{Material}}} &
  \multicolumn{1}{c}{\multirow{2}{*}{\textbf{Image}}} &
  \multicolumn{5}{c}{\textbf{\% by mass}} \\ 
\multicolumn{1}{c}{} &
  \multicolumn{1}{c}{} &
  \multicolumn{1}{c}{Linen} &
  \multicolumn{1}{c}{Viscose} &
  Cotton &
  Wool &
  Polyester \\ \midrule\hline
Cotton-Linen  & \includegraphics[width=.1\linewidth,valign=m]{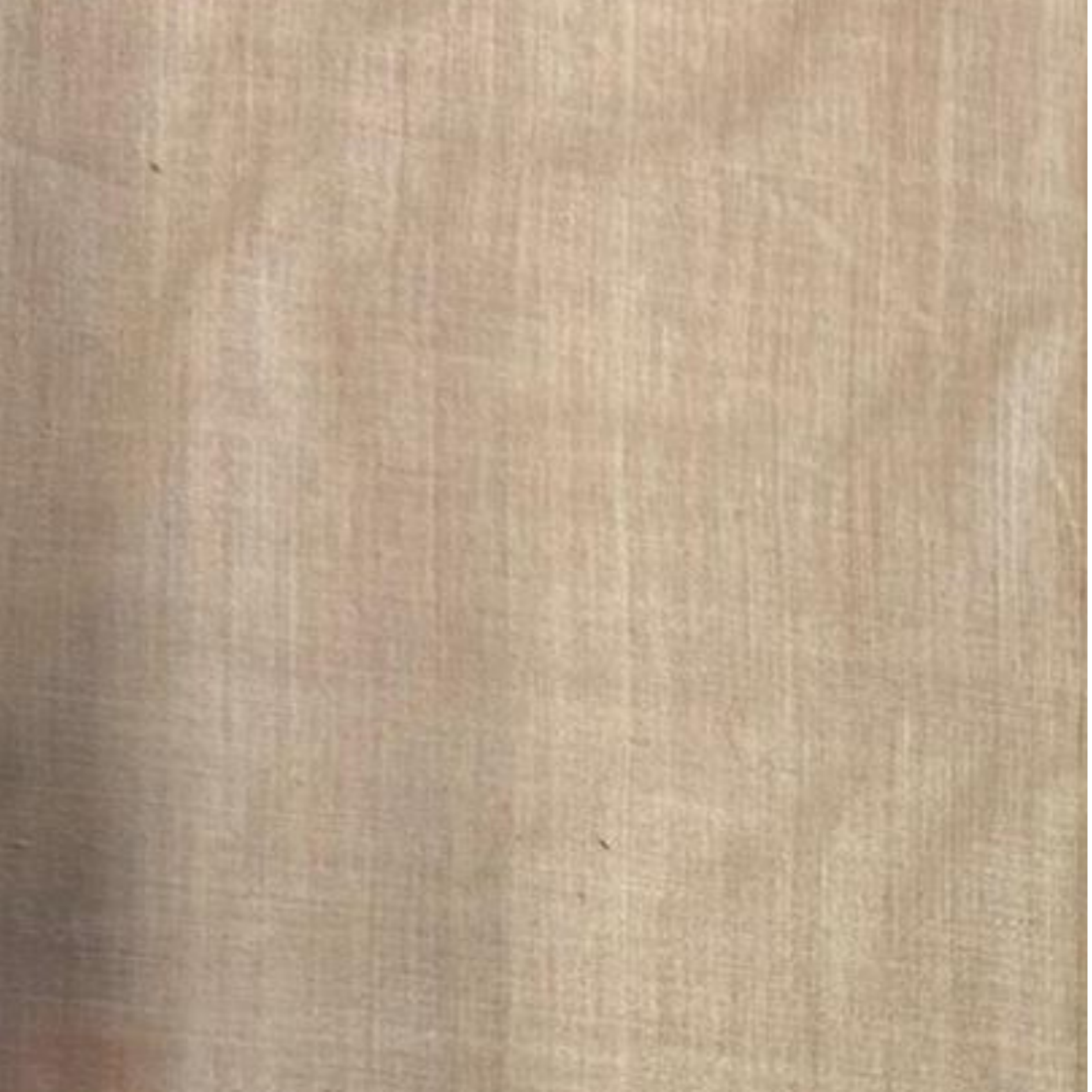} & 45 & 0    & 55 & 0    & 0 \\
Poplin        & \includegraphics[width=.1\linewidth,valign=m]{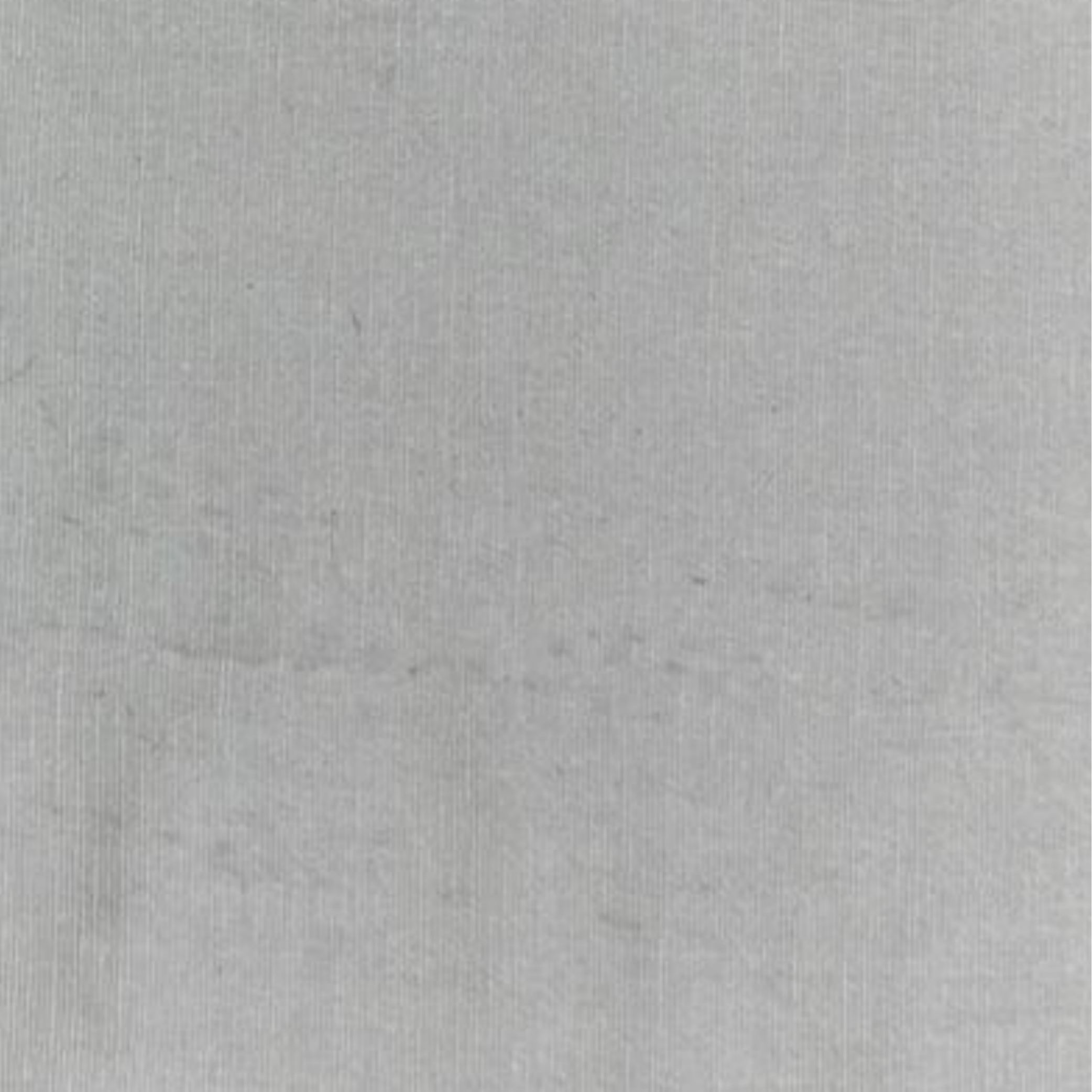} & 0    & 0    & 20  & 0    & 80 \\
Drill Stretch & \includegraphics[width=.1\linewidth,valign=m]{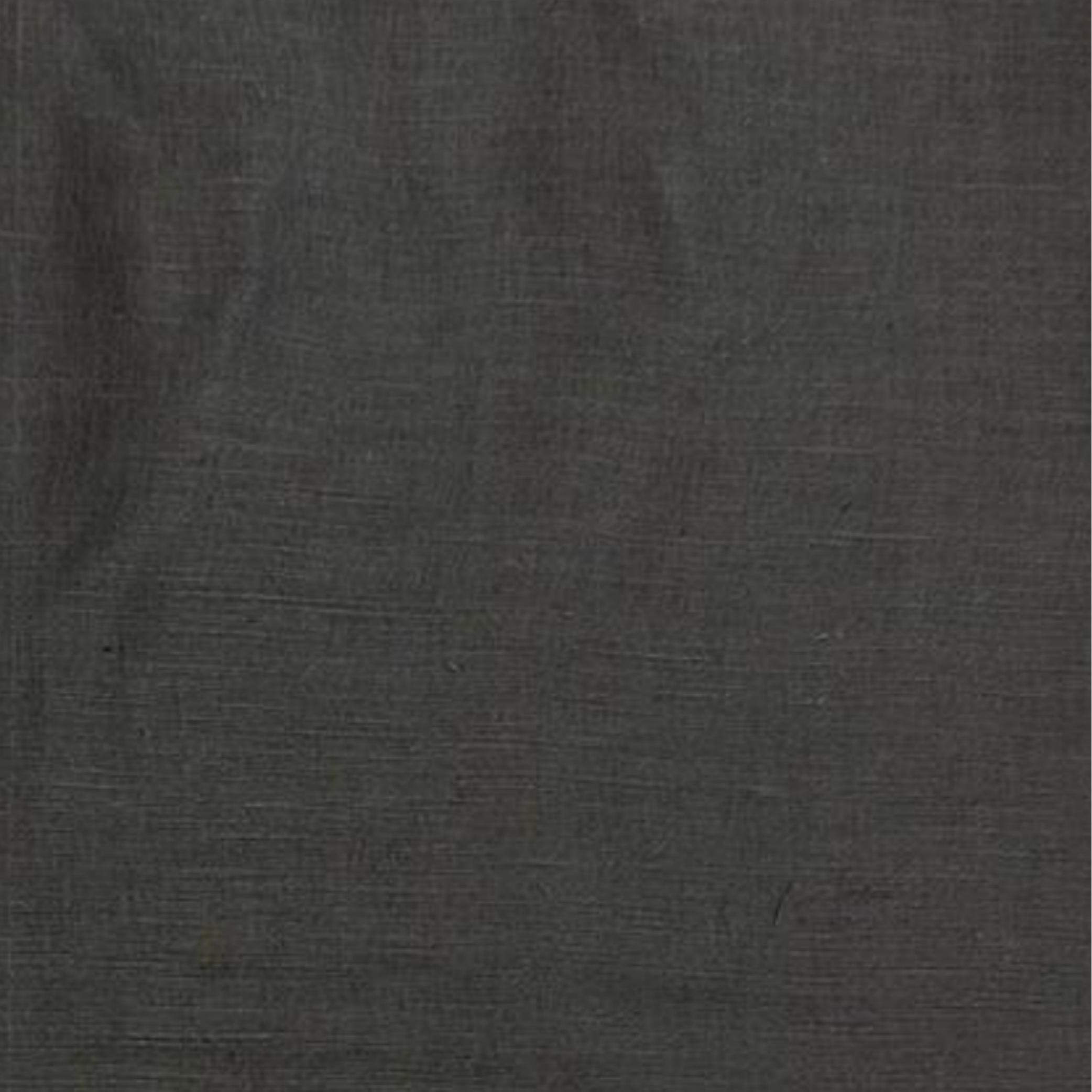} & 0    & 0    & 100    & 0    & 0 \\
Felt          & \includegraphics[width=.1\linewidth,valign=m]{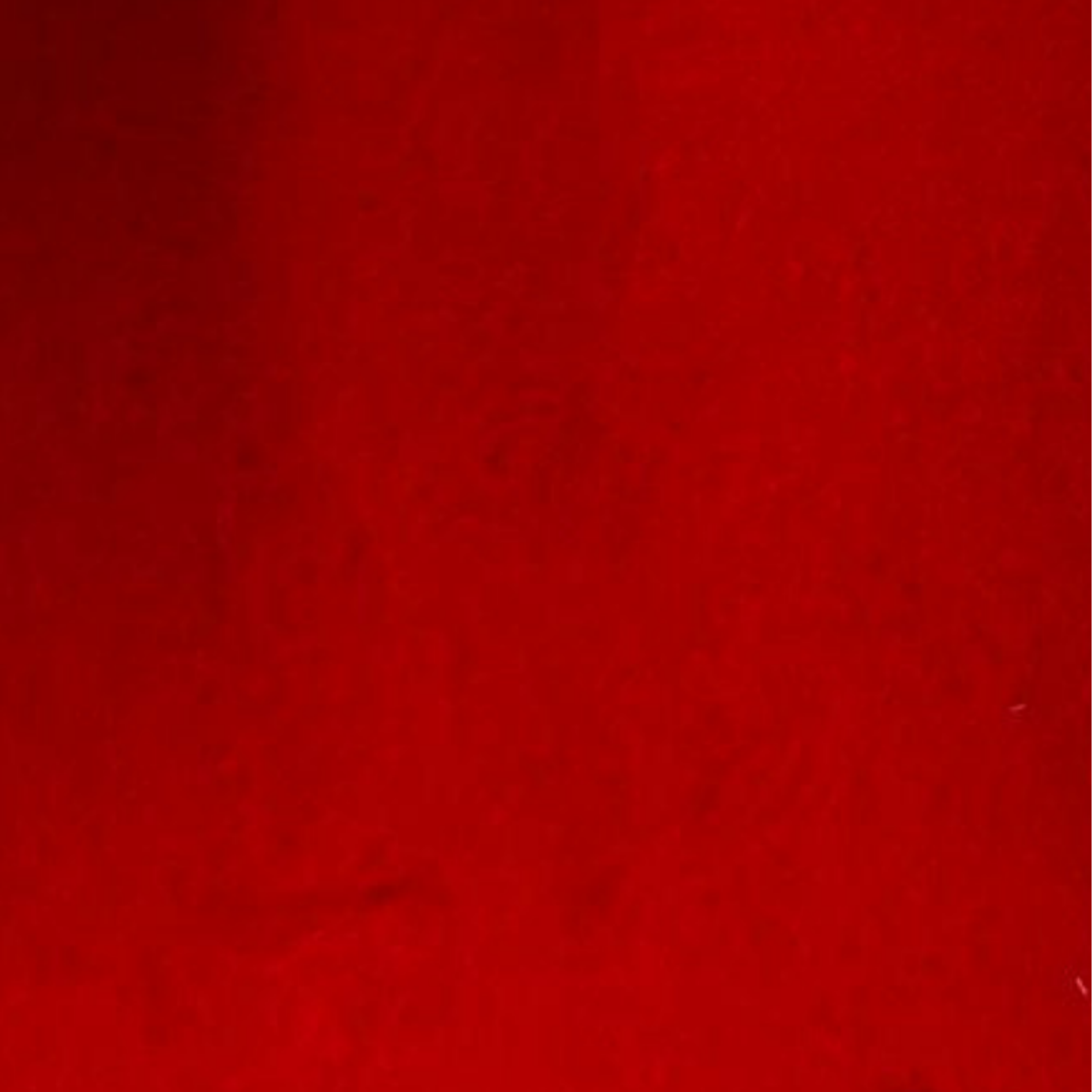} & 0    & 65 & 0    & 35 & 0 \\ \hline
\end{tabular}
}
\end{table}

\section{Method}
\label{sec:method}
We present a unified approach to tackle heterogeneous sensors and efficient knowledge transfer in TRL. Our approach relies on a unifying format for different sensor data, and exploits convolution's agnosticiity to input size to enable fully shared models. These fully shared models in turn enables convenient knowledge transfer. We also discuss data augmentations to counteract limited tactile training data. Lastly, we discuss a continual tactile learning approach as a direct application of knowledge transfer.

\subsection{Tactile Images and Convolutional Architectures}
\label{sec:method_tactile_image}
We use simple transformations to convert data generated by various sensors into 2D images, which serves as the unified input format for the subsequent ML models. Specifically, tactile images aims to transform tactile sensory output into an encoding of the global geometry for the contact surface. This transformation is inspired by the processing similarities between the human visual cortex and somatosensory system~\cite{doi:https://doi.org/10.1002/9781118133880.hop203008}, and captures the intuition that significant tactile properties are fundamentally spatial~\cite{Haindl2013VisualTA}.

\begin{wrapfigure}{r}{0.4\linewidth}
    \includegraphics[width=\linewidth]{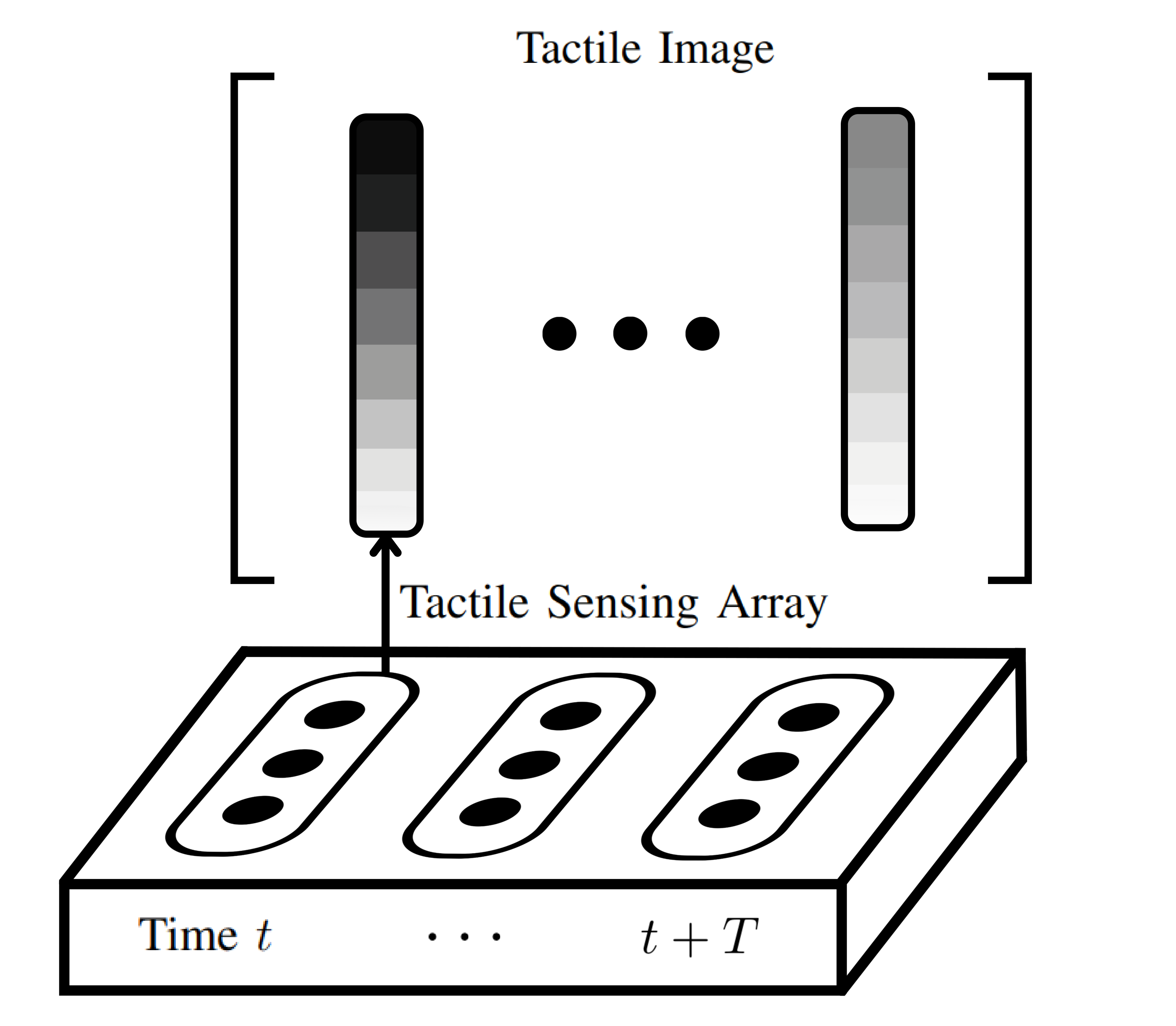}
    \caption{Tactile Image processing for non-camera-based sensors.}
\label{fig:tacim_gen}
\end{wrapfigure}

Camera-based sensors such as GelSight directly capture \textit{global} surface geometry as images and can be used as model input directly. However, non-camera-based sensors typically have sparse sensing points that only produce \textit{localized} signals about the contact surface. To better encode the global surface geometry, we thus require more local samples that span across the contact surface. This could be conveniently achieved by concatenating consecutive vectors from the tactile data stream, as the sensor slides over the contact surface. Formally, let $S=\{s_1, s_2, \dots, s_T\}$ be the data stream produced by a sensor sliding across a surface, where $s_t\in\mathbb{R}^n$ is a single reading from the sensor. We define a tactile image as a matrix $\textrm{Im}(S) = [s_j, s_{j+1}, \dots, s_{k}]$ for some constant $j, k$. Intuitively, $Im(S)$ leverage the temporal dimension of tactile data stream to better encode global surface properties (see also \cref{fig:tacim_gen} for an illustration).

We note that tactile images of different sensors still have different dimensions. To achieve fully shared models for knowledge transfer, we thus adopt convolutional architectures such as ResNet~\cite{He_2016_CVPR}, since convolution does not require a fixed input size. ResNet is also a representative state-of-the-art model for processing spatial input, including the surface geometry encoded in tactile images. 

\begin{figure}[h]
      \centering
      \includegraphics[width=\linewidth]{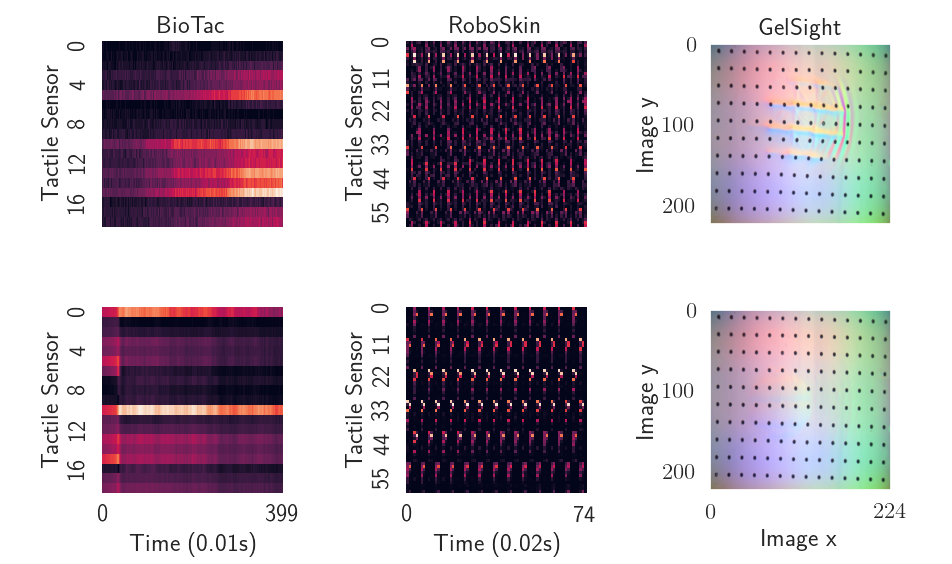}
      \caption{Tactile image representations for the BioTac, RoboSkin and GelSight sensors for two material classes.}
      \label{fig:tacims}
\end{figure}

\subsection{Model Training}
\label{sec:method_train}
With tactile images and our chosen model architecture, we effectively recast TRL as a vision task. For training, we minimise the empirical cross-entropy loss
\begin{equation}
    \argmin_{f} \sum_{(x, y) \in \mathrm{D}} \ell_{ce}(f(x), y)
\end{equation}
where $f$ is the model and $\ell_{ce}$ is the cross-entropy loss. $\mathrm{D}$ denotes the dataset containing labeled tactile images $(x, y)$.

Crucially, we can initialize $f$ with a pre-trained model to enable knowledge transfer. In particular, we may interpret TRL as a downstream task for a vision foundational model on general spatial features. In our experiments, we will demonstrate that a foundational model trained on natural images already robustly encodes the general features required for tactile images.

 \paragraph{Data Augmentation} As discussed earlier, tactile datasets are typically small due to the high cost of data collection due the interactivity of the modality and significant wear and tear. Data augmentation is therefore important to mitigate model overfitting, especially for larger architectures like ResNet. We propose to directly apply standard CV augmentations: \textit{resizing, cropping, flipping} and \textit{jittering}. We observe that each of these augmentations encodes a meaningful variation to the data collection process, even for non-camera-based sensors. For instance, cropping the tactile images encodes varying the duration of robot motion during data collection. \cref{tab:tactile_aug} lists all chosen augmentations and their interpretation.

\begin{table}[th]
    \centering
    \caption{Tactile images augmentations and their physical interpretation}
    \begin{tabular}{ll}
    \midrule
         \textbf{Augmentation Technique} & \textbf{Physical Interpretation}\\
         \midrule\midrule
         Flipping (along data axis) & Reversing the direction of robot motion.\\
         Resizing (along temporal axis) & Vary the speed of robot motion.\\
         Cropping (along temporal axis) & Vary the duration of robot motion.\\
         Jittering & Simulate sensor noise and drift.\\ \midrule
    \end{tabular}
    \vspace{-1em}
    \label{tab:tactile_aug}
\end{table}

The chosen augmentations are readily accessible from common deep learning frameworks~\cite{paszke2017automatic} and may be directly applied. We will demonstrate empirically that the augmentations is crucial to model robustness.

\subsection{Continual Tactile Learning}
As robots are increasingly expected to work in unstructured environments, continual learning of unordered new percepts is important. \cref{sec:bg_tasks} introduced continual learning (CL) of new materials as a natural extension to standard material classification. 
The two key challenges for CL are: 1) whether robots could learn about new materials on the fly, and 2) continuous learning does not cause catastrophic forgetting of current knowledge~\citep{mccloskey1989catastrophic, ratcliff1990connectionist}.

We adopt schedule-robust online continual learning (SCROLL)~\cite{wang2022schedule} to tackle CL of new materials. We choose SCROLL because the method leverages pre-trained models for efficient knowledge transfer, thus allowing new materials to be learned with limited interaction. In addition, SCROLL is robust to the schedule under which the data is presented (e.g., the order in which each material is learned), a crucial property to ensure model reliability in real-world situations.

Using the notations introduced in Eq. \cref{eq:cl}, we characterize SCROLL as a two-phase process. Given a suitable pre-trained embedding model $\embd$, we first learn an online linear classifier $\ridge_t$ via recursive least squares~\citep{kailath2000linear} as novel material data $B_t$ is observed. We then fine-tune the composite model $f_t = \embd \circ \ridge_t$ using the current memory buffer $M_t$ to yield $f_t^*$. Both $f_t$ and $f_t^*$ are valid CL models for all data observed so far, with $f_t^*$ having a fine-tuned representation based on the observed data.  SCROLL uses exemplar selection~\cite{rebuffi2017icarl} for updating $M_t$. The overall algorithm is presented in \cref{alg:scroll},

\begin{algorithm}[h]
\caption{SCROLL (incremental)\label{alg:scroll}}
    \begin{algorithmic}
        \small
        \STATE {\bfseries Initialization:} Buffer $M_0 = \varnothing$, data statistics $c_y^0=0, A_0=0$
        \vspace{0.2em}
        \STATE {\bfseries Input:} Embedding model $\embd$, next data batch $B_t$, current buffer $M_{t-1}$, current data statistics $c_y^{t-1}, A_{t-1}$
        \vspace{0.5em}
        \STATE $c_y^t, A_t = \textrm{RecursiveLeastSquare}(c_y^{t-1}, A_{t-1})$
        \STATE $\ridge_t = \textrm{RidgeRegressor}(c_y^t, A_t)$
        \STATE $f_t=\ridge_t\circ\embd$
        \STATE $M_t = \textrm{SelectExemplar}(M_{t-1}, B_t, \embd)$
        \STATE  $f_t^*$ = \textrm{FineTune}$(f_t, M_t)$
        \STATE {\bfseries Return} $c_y^t, A_t, M_t, f_t$ and $f_t^*$
    \end{algorithmic}
\end{algorithm}
\noindent where $c_y, A$ are necessary data statistics for recursive least squares (see \cite{wang2022schedule} for further details on SCROLL).

\section{Experiments}
We evaluate our approach  extensively across a wide variety of sensors and tasks, as introduced in \cref{sec:related_work,sec:sensor_data}. Our experiments address the following questions:
\begin{itemize}
    \item Is our approach generally applicable to heterogeneous tasks and sensors? How does our approach compared to sensor-specific methods?
    \item What are the effects of tactile image augmentation?
    \item Does our approach allow efficient knowledge transfer? What are the effects of knowledge transfer?
\end{itemize}


\paragraph{Data Pre-Processing} Following \cref{sec:method_tactile_image}, we transform BioTac data into $19 \times 400$ images by stacking 400 consecutive vectors. This corresponds to 4 seconds of data. RoboSkin data is transformed into $60 \times 75$ images, corresponding to 1.5 seconds of data. Lastly, Contactile data is transformed into $27 \times 599$ images, which is 6 seconds of data. We note that the exact size for the temporal dimension is not crucial, since we will also leverage random cropping and resizing along the temporal dimension for data augmentation. Since these tactile images only have a single channel, the channel is repeated three times to match the input dimension for the vision foundational model used in the experiments. All tactile images and GelSight data is normalized to the range of $[-1, 1]$.

\paragraph{Model Architecture and Pre-training} We choose a ResNet-18 pre-trained on MetaDataset~\cite{Triantafillou2020Meta-Dataset:} as our foundational vision model. It is chosen for its balanced accuracy and computational efficiency. We emphasize that other foundational models may be easily chosen given the trade-off between accuracy and efficiency. We also highlight all experiments use the \textit{identical} foundational model without any modification, as our approach allows fully shared models.

\subsection{Standard Material Classification} \label{sec:mat_class}
We compare our approach with baseline methods on standard material classification using BioTac-20, RoboSkin and GelSight datasets. We highlight that the baselines are specifically tailored to the BioTac or RoboSkin sensors, whilst our model is generic.

\paragraph{Model Details} Our model is trained for 100 epochs using stochastic gradient descent (SGD). A validation set is employed to schedule the learning rate, mitigating performance plateaus. An initial learning rate of $0.01$ is chosen empirically, with a momentum of $0.9$ and a weight decay of $0.0001$. 5-fold cross validation is performed for all experiments.

\paragraph{Baseline Methods} We compare our approach to a diverse set of methods investigated in \cite{9340693}, including a spiking neural network (SNN), LSTM, regular support vector machine (SVM) and spike-encoded SVM (SVM Spike).

Table~\ref{tab:supervised} reports the classification accuracy for all evaluated methods. Our generic ResNet outperforms the baselines by more than 4\%, suggesting the viability of our tactile image approach. In addition, the results clearly shows that fine-tuning from the foundational model is more advantageous than random initialization. This indicates positive knowledge transfer from the pre-trained model and improved generalization. This is especially visible for the GelSight dataset owing to the imbalance between the small size of the dataset and the large input dimension. 

\begin{figure}[h]
      \centering
      \includegraphics[width=0.8\linewidth]{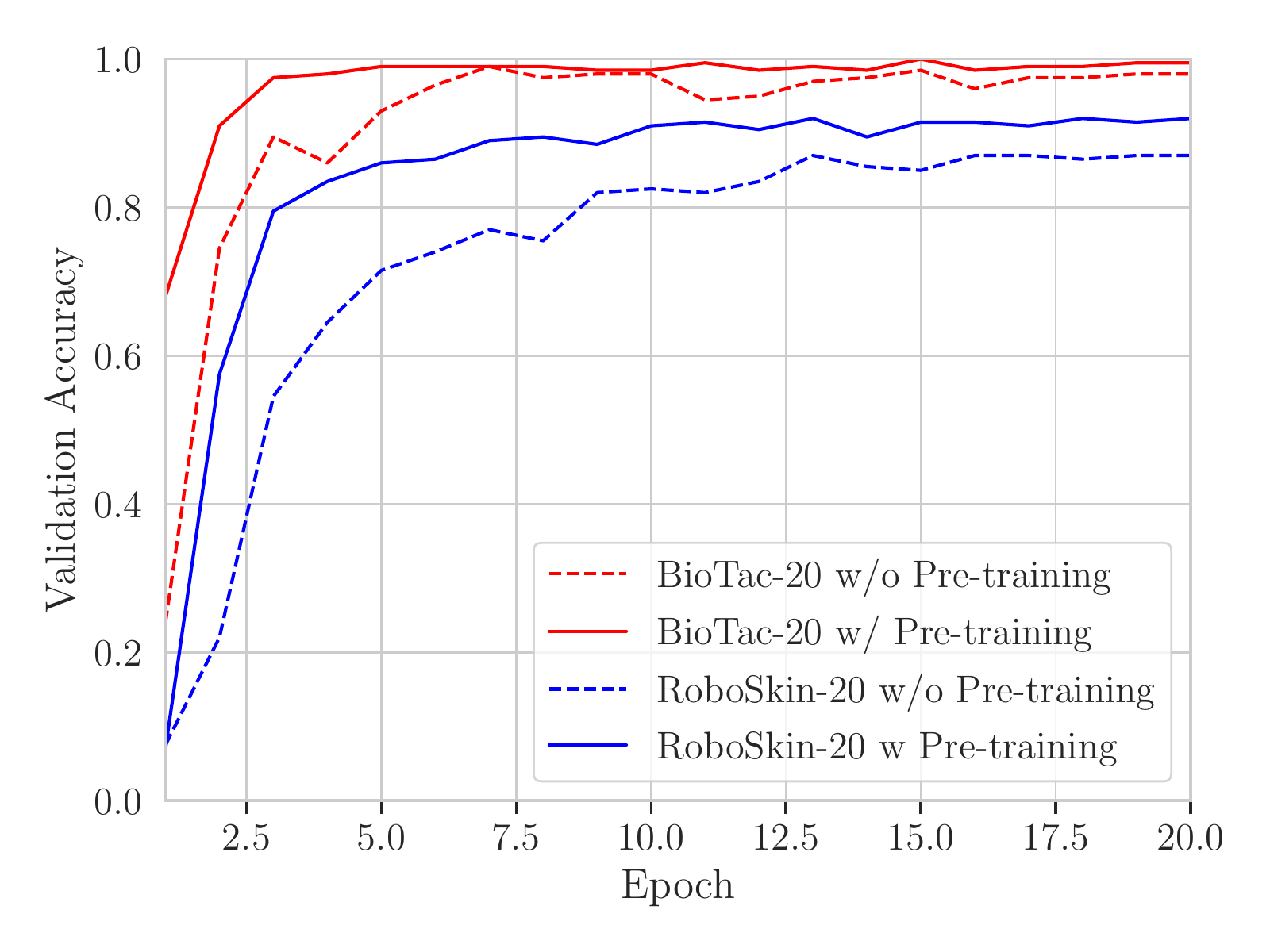}      \caption{Test Accuracy over the first 20 epochs for both BioTac-20 (red) and RoboSkin 20 (blue), with (solid) and without (dashed) pre-training.}
      \label{fig:learning_efficiency}
\end{figure}

Pre-training also noticeably improves learning efficiency, as reported in \cref{fig:learning_efficiency}. For both BioTac-20 and RoboSkin datasets, transferring from the foundational model (i.e., with pre-training) achieves higher accuracy with fewer iterations over the training data. Learning efficiency is a desirable property for robots requiring fast adaptation to novel environments.

\begin{table}[t]
\centering
\caption{Material Classification Accuracy (\%). Numbers for baseline methods are originally reported in \cite{9340693}. Pre-train denotes initialization with the foundational vision model.}
\label{tab:supervised}
\footnotesize
\begin{tabular}{lccc}
\midrule
\textbf{Method} & \textbf{BioTac-20} & \textbf{RoboSkin} & \textbf{GelSight} \\ \midrule\midrule
SVM             & $94.2 \pm 0.7$        & $50.5 \pm 5.6$   &  n.a   \\
SVM (spikes)    & $93.5 \pm 1.5$         & $63.3 \pm 1.8$   &  n.a  \\
Conv-LSTM       & $94.5 \pm 1.5 $        & $93.5 \pm 0.5$    &  n.a \\
SNN             & $94.6 \pm 1.3$         & $92.2 \pm 0.5$    &  n.a  \\
\hdashline[1pt/1pt]\noalign{\vskip 1.0ex}
Least Square w/ Pre-train & $93.8 \pm 1.2$  & $84.8 \pm 1.3$  & $67.1 \pm 0.8$       \\

ResNet (ours) & $98.0 \pm 0.3$ & $95.0 \pm 0.6$  & $92.9 \pm 0.3$ \\
ResNet w/ Pre-train (ours)    & $\mathbf{98.9 \pm 0.2}$         & $\mathbf{96.0 \pm 0.5}$   &   $\mathbf{95.1 \pm 0.3}$     \\ \midrule
\end{tabular}
\end{table}

\paragraph{Foundational Models and Tactile Images} To better understand the connection between our foundational model and tactile images, we introduce another baseline in \cref{tab:supervised} denoted by ``Least Square''. This baseline encodes all tactile images into fixed representations using the pre-trained ResNet, and only learns a least-squares classifier over the fixed representation. The accuracy of this baseline thus directly reflects the usefulness of the pre-trained model towards tactile images. Surprisingly, the results show that the foundational vision model trained from natural images already contains the general features required for tactile texture representation, despite the apparent distributional shift. This provides direct support to the connection between visual and tactile perception, resembling the similarities between the human visual cortex and somatosensory system. The results also provide empirical justification for our choice of tactile images as model input.

\subsection{Augmentation and Model Robustness}
\label{sec:augs}
As noted in \cref{sec:method_train}, data augmentations applied to tactile images may be interpreted as diversifying the conditions of data collection. This is crucial for tactile datasets as they are generally expensive to collect. We investigate the effects of augmentation in the following experiments.

\begin{figure}[h]
     \centering
     \begin{subfigure}[b]{0.49\linewidth}
         \centering
         \includegraphics[width=\linewidth]{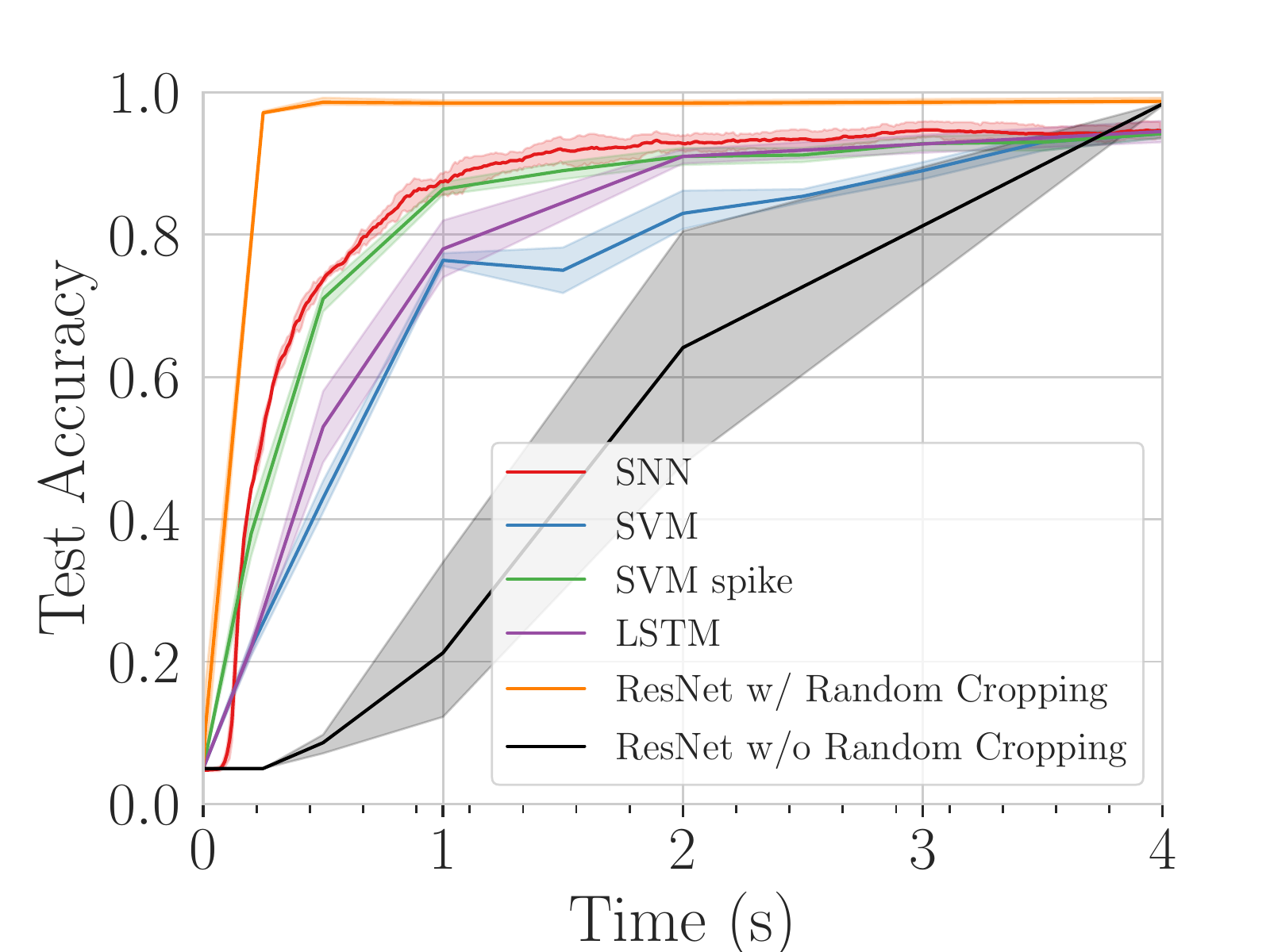}
         \caption{BioTac-20}
         \label{fig:biotac50_len}
     \end{subfigure}
     \hfill
     \begin{subfigure}[b]{0.49\linewidth}
         \centering
         \includegraphics[width=\linewidth]{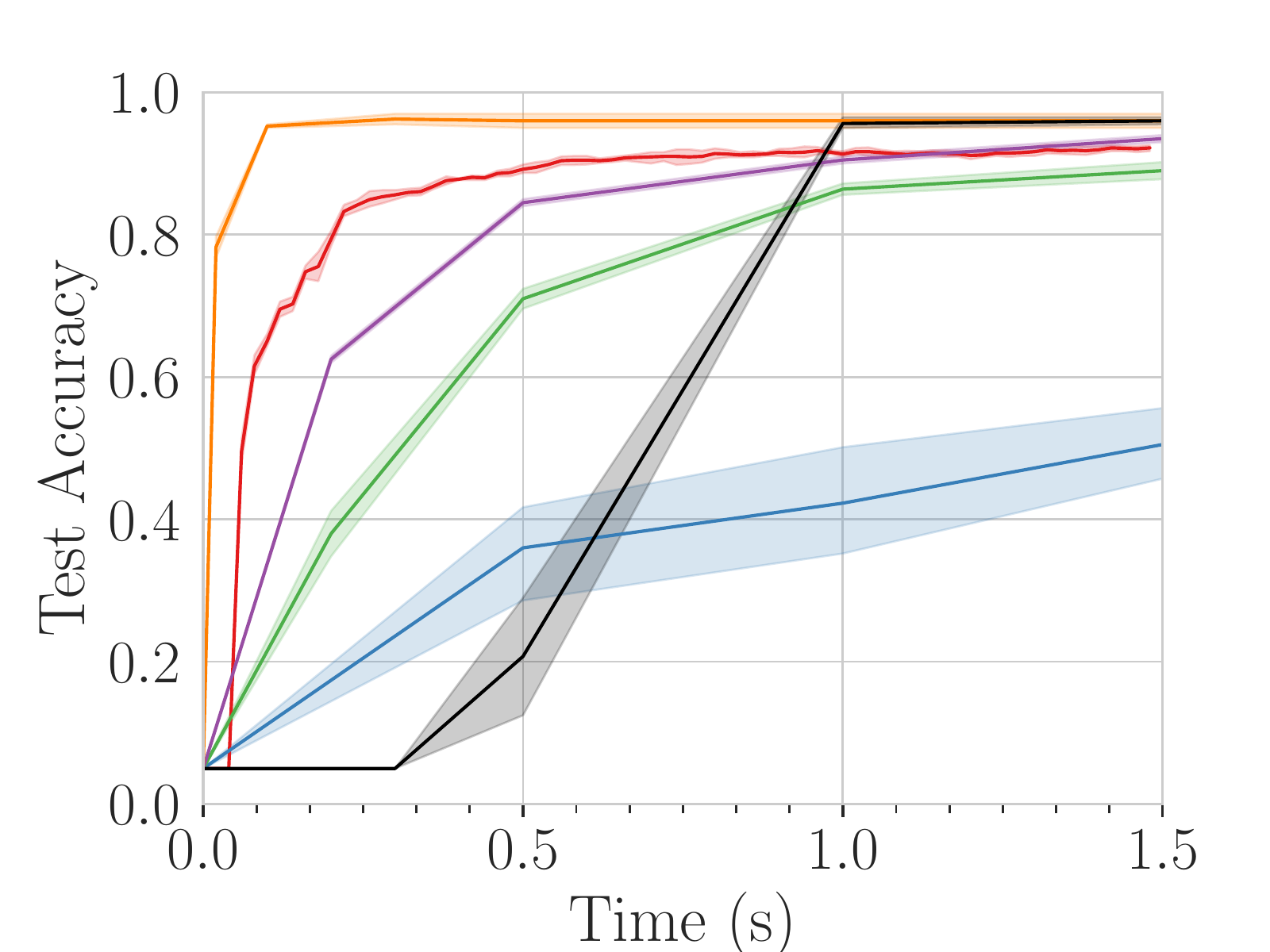}
         \caption{RoboSkin-20}
         \label{fig:roboskin20_len}
     \end{subfigure}
        \caption{Test accuracies of our approach with and without random cropping augmentations for varying data length. Baseline methods included for comparison.}
        \label{fig:random_length}
\end{figure}

\paragraph{Robustness to Sampling Length} For material classification, it is desirable to shorten the sampling length without sacrifice to accuracy. This corresponds to classifying randomly cropped tactile images in our formulation. It was also investigated in \cite{9340693} as a strength of spiking neural architecture. In \cref{fig:random_length}, we investigate how random cropping affects classification accuracy over varying data length, and compare our approach to previous methods.

The results clearly show that our model with augmentation outperforms the previous methods, achieving higher test accuracy with less data required. For both datasets, ResNet with augmentation is able to accurately classify the materials with about 0.3 seconds of sensor data. As the data length increases, the test accuracy rapidly increases and remains high, suggesting that our model could efficiently accumulate information over short duration while maintaining robustness over long run. In addition, \cref{fig:random_length} shows that augmentation is crucial for robust performance. The same model trained without augmentation performed the worst among all methods, suggesting overfitting to the original data length and less robust features learned.

\paragraph{Robustness to Movement Speed} While some tactile datasets are collected under a tightly controlled robot motion, it is preferable that the learned model generalizes to more varied motions. We simulate different speeds of the robot's sliding motion during tactile sensing by sub-sampling the test set data along the temporal axis, and investigate the effects of augmentation on this out-of-distribution test set.

\begin{figure}[h]
     \centering
     \begin{subfigure}[b]{0.49\linewidth}
         \centering
         \includegraphics[width=\linewidth]{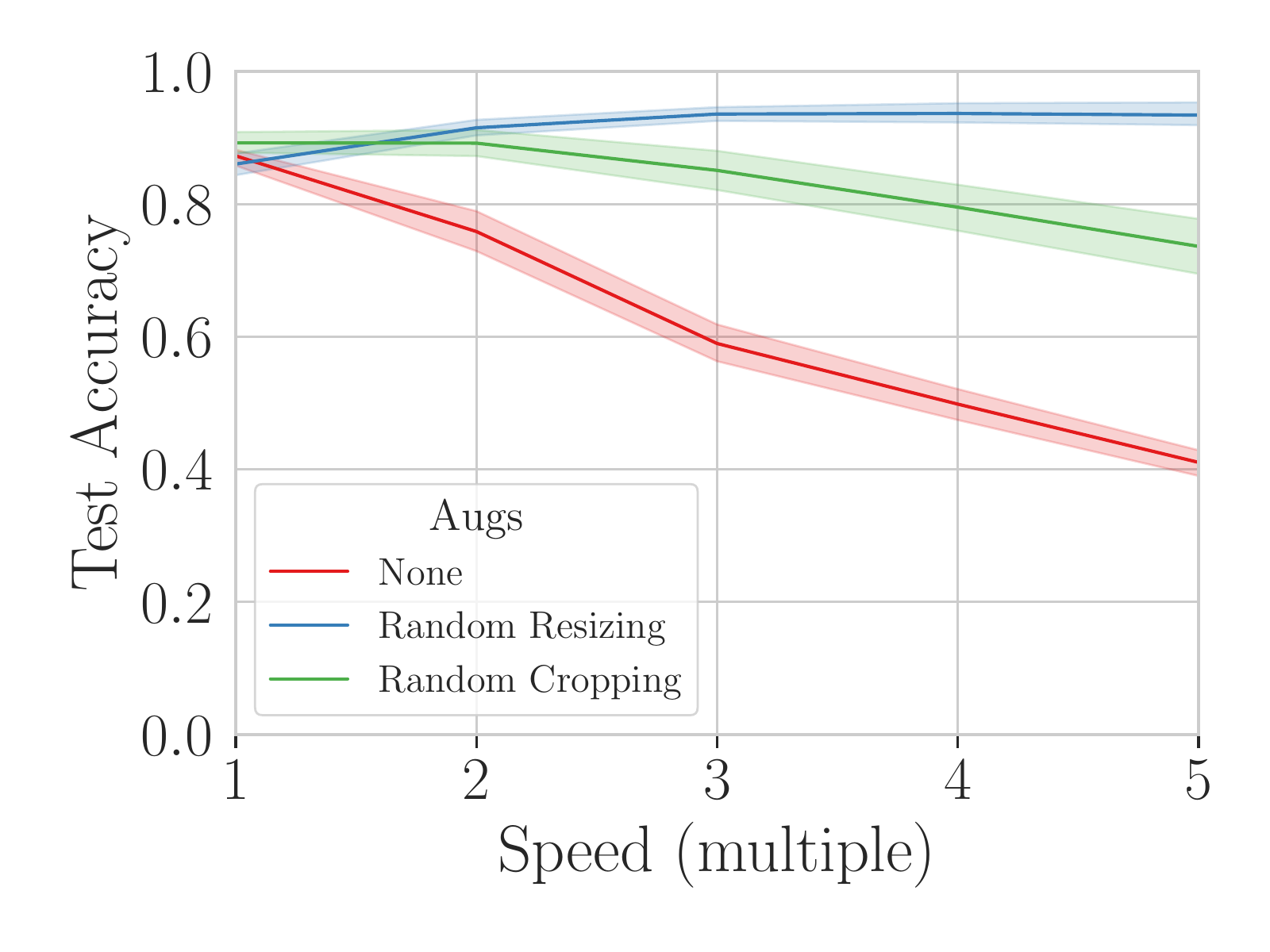}
         \caption{BioTac-50: Speed}
         \label{fig:biotac50_rate}
     \end{subfigure}
     \hfill
     \begin{subfigure}[b]{0.49\linewidth}
         \centering
         \includegraphics[width=\linewidth]{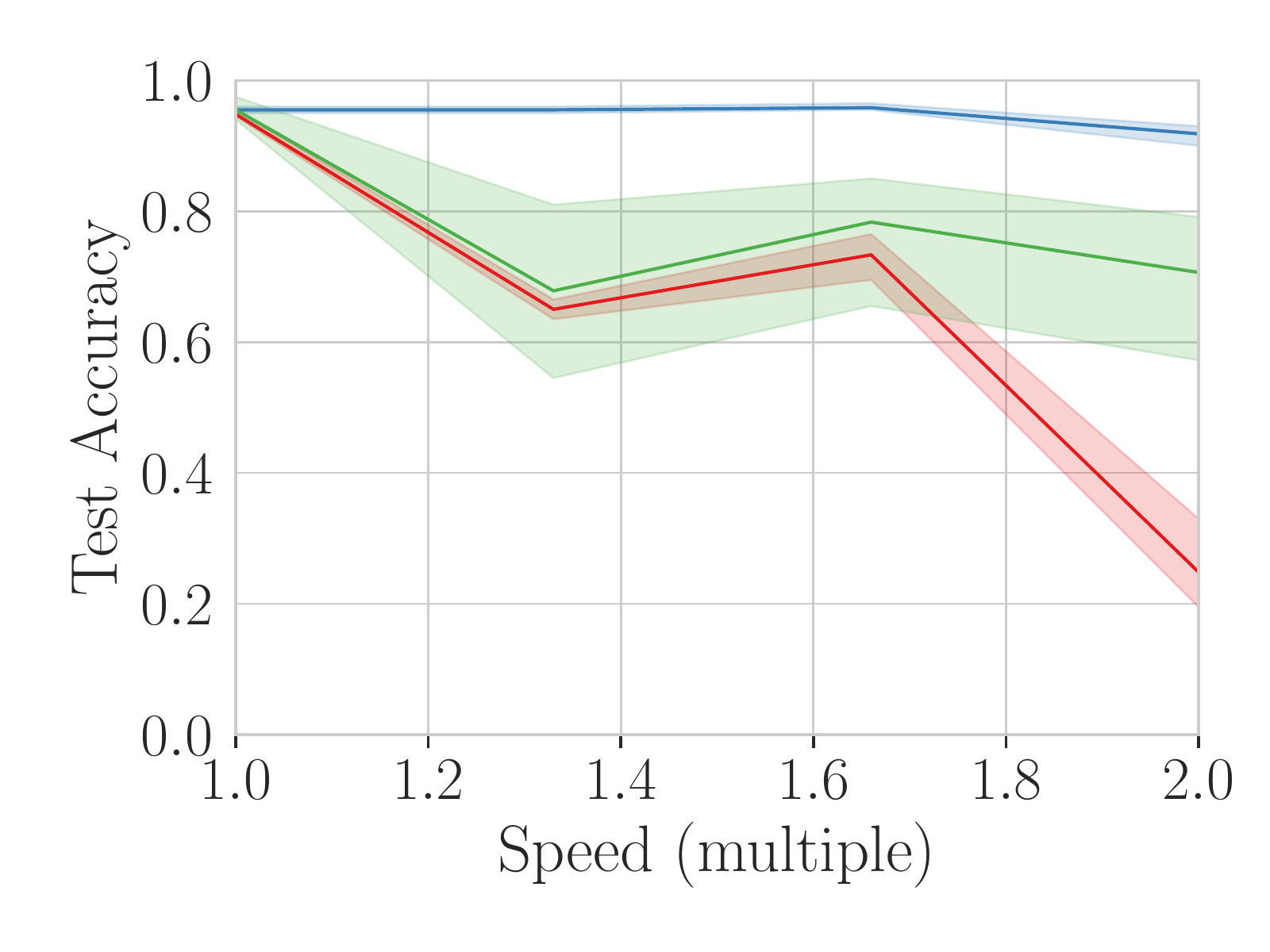}
         \caption{RoboSkin-20: Speed}
         \label{fig:roboskin20_rate}
     \end{subfigure}
     \qquad
        \caption{The effects of augmentation with respect to varying robot movement speed during tactile sensing. X-axis denotes the multiples of the original robot speed.}
        \label{fig:random_rate_dir}
\end{figure}

\cref{fig:random_rate_dir} shows that the model trained with random resizing augmentation is robust against varying robot speed, achieving consistent accuracy across different movement speed. In contrast, the model with no augmentation generalized poorly even with slight speed deviation. The figure also shows that random cropping improves model robustness against varying movement speed.

\begin{figure}[h]
     \centering
    \begin{subfigure}[b]{0.49\linewidth}
         \centering
         \includegraphics[width=\linewidth]{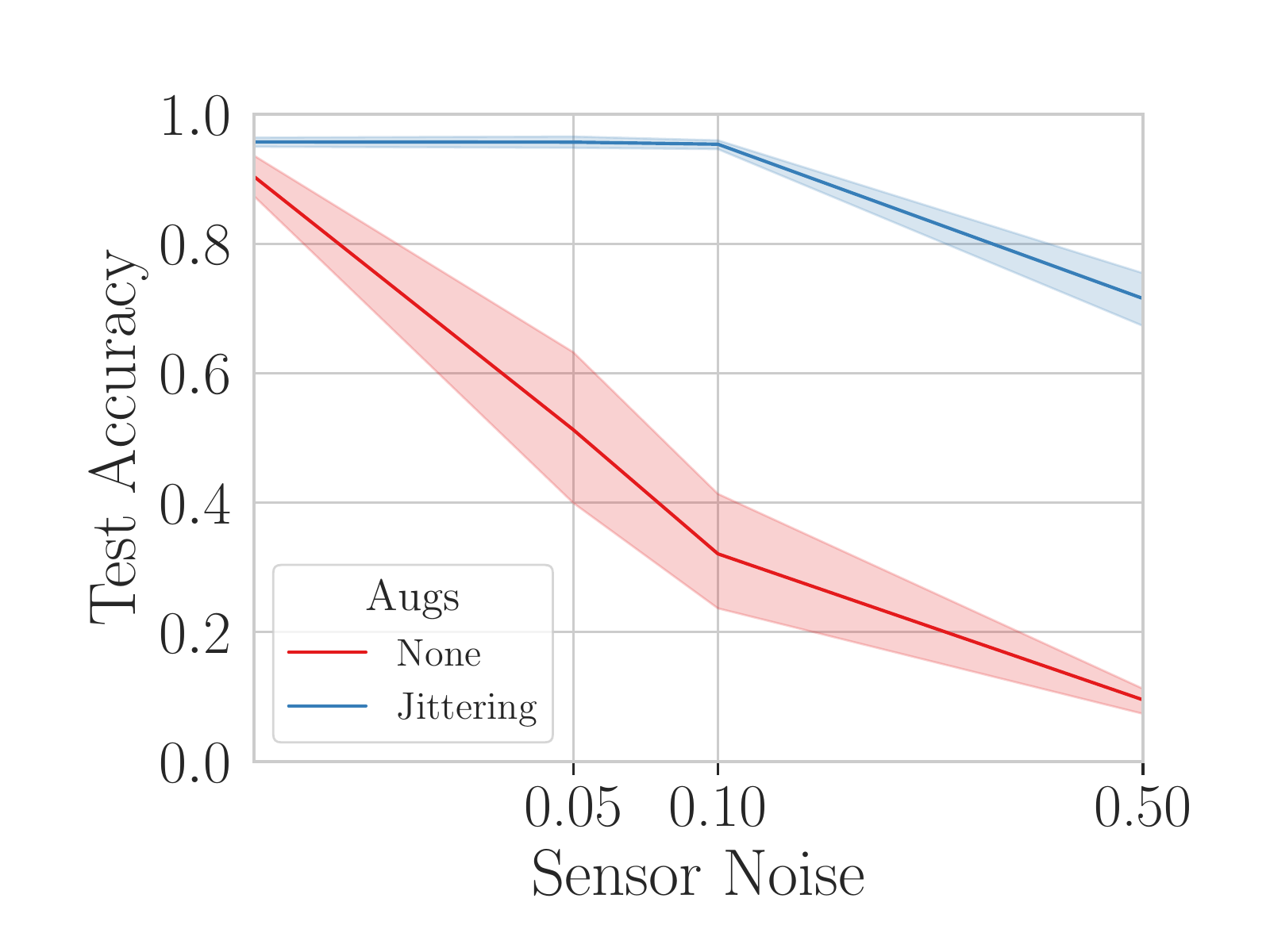}
         \caption{BioTac-50: Noise}
         \label{fig:biotac50_force}
     \end{subfigure}
     \hfill
     \begin{subfigure}[b]{0.49\linewidth}
         \centering
         \includegraphics[width=\linewidth]{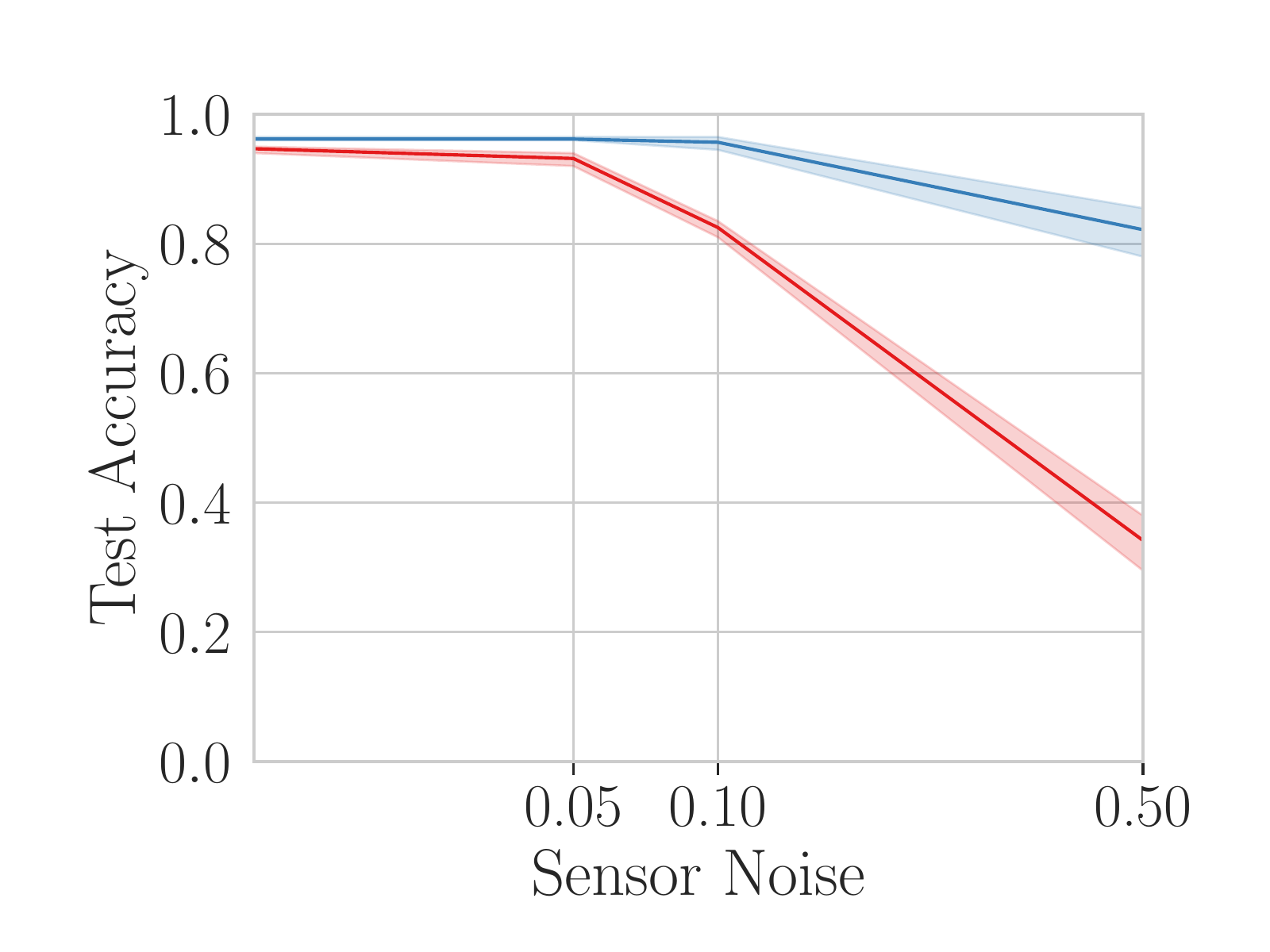}
         \caption{RoboSkin-20: Noise}
         \label{fig:roboskin20_force}
     \end{subfigure}
        \caption{The effect on test accuracy with respect to sensor noise. X-axis denotes maximum noise level added to tactile images.}
        \label{fig:random_noise_dir}
\end{figure}

\paragraph{Robustness to Sensor Noise} Similar to the previous experiment, we construct another out-of-distribution test set by injecting random sensor noise. \cref{fig:random_noise_dir} and evaluates the effects of augmentations.

\begin{figure*}[!h]
     \centering
     \begin{subfigure}[b]{0.32\linewidth}
         \centering
         \includegraphics[width=\linewidth]{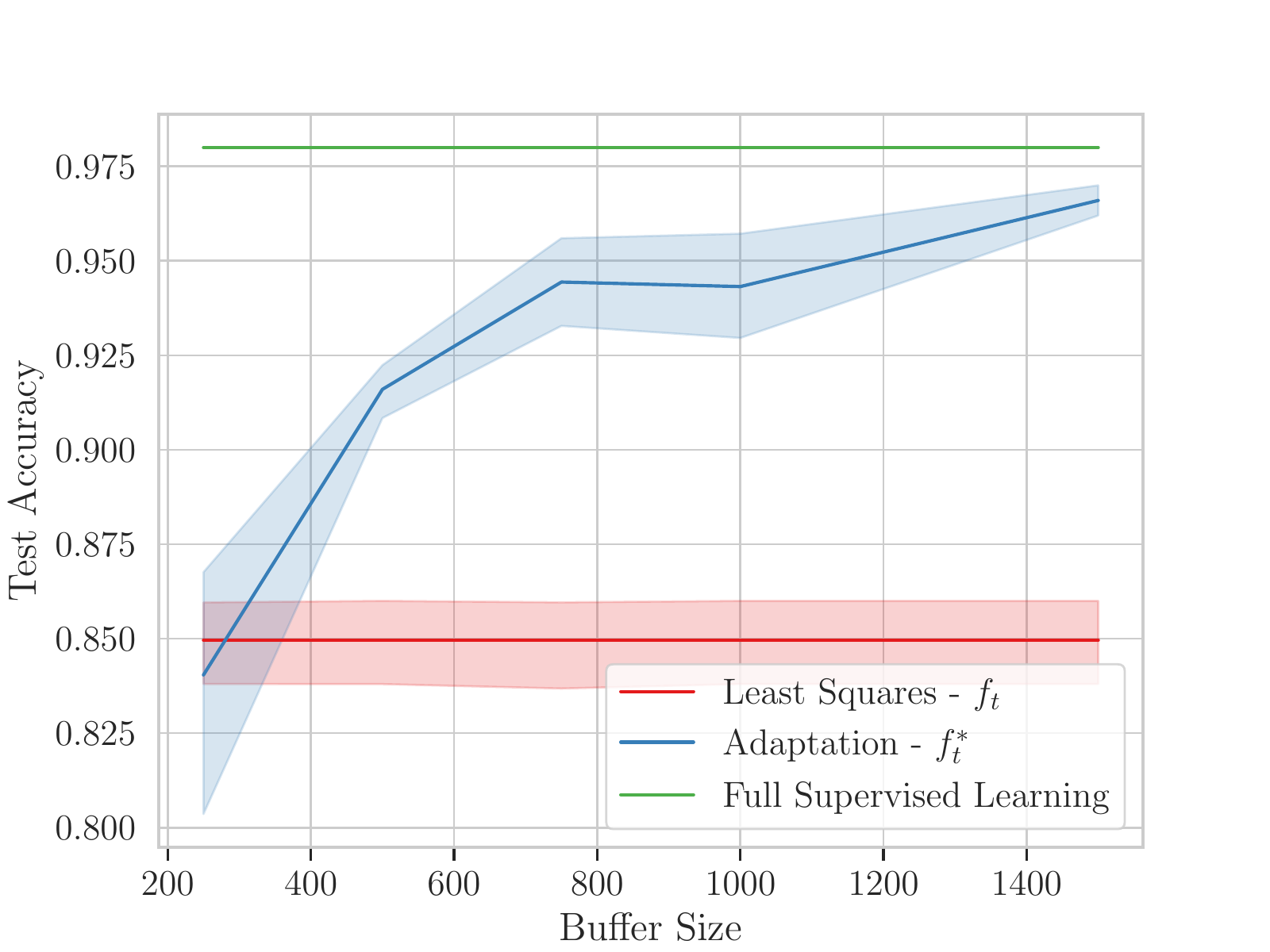}
         \caption{BioTac-50}
         \label{fig:bio50_cl_buff}
     \end{subfigure}
     \hfill
     \begin{subfigure}[b]{0.32\linewidth}
         \centering
         \includegraphics[width=\linewidth]{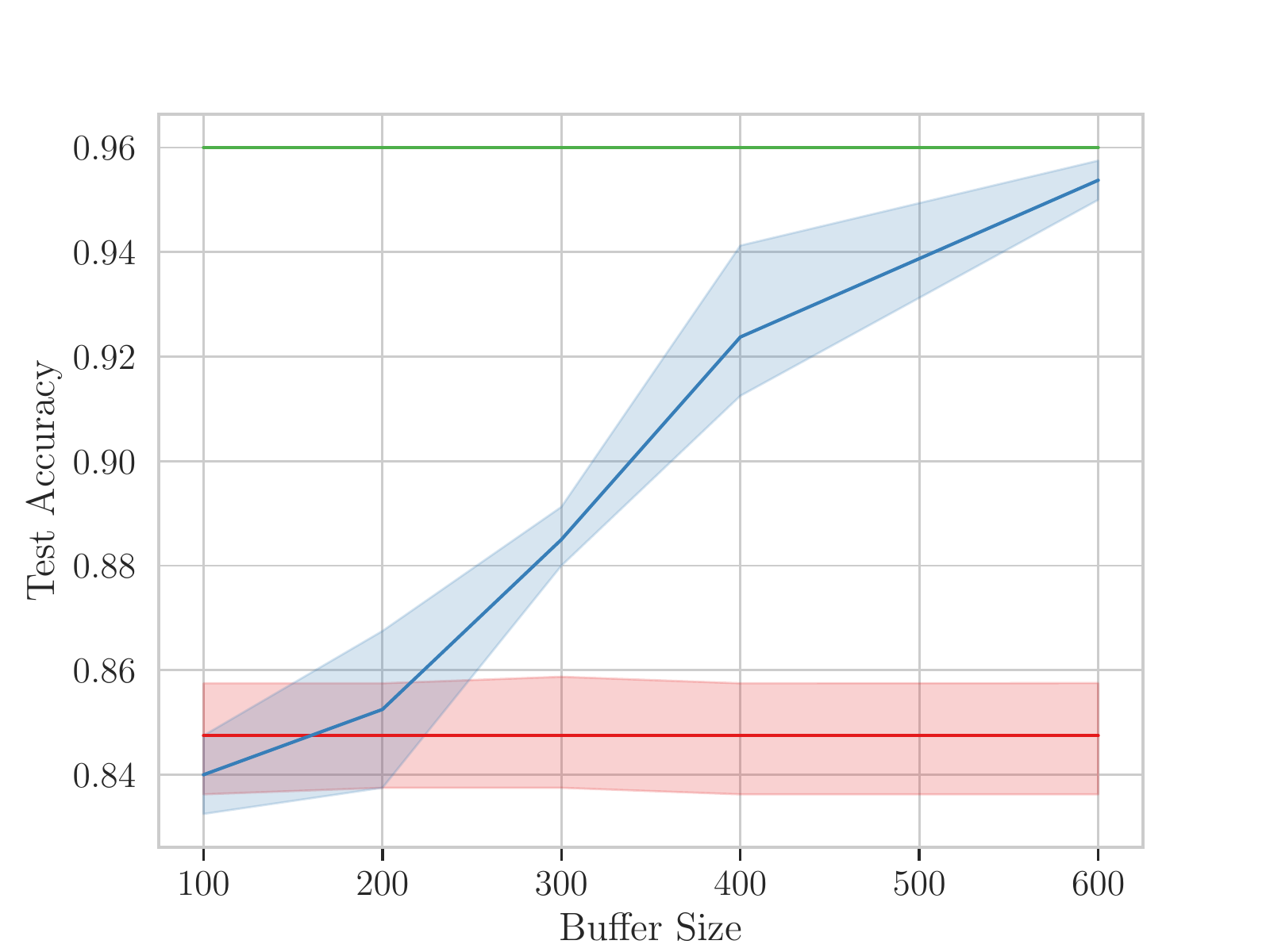}
         \caption{RoboSkin-20}
         \label{fig:icub20_cl_buff}
     \end{subfigure}
     \hfill
     \begin{subfigure}[b]{0.32\linewidth}
         \centering
         \includegraphics[width=\linewidth]{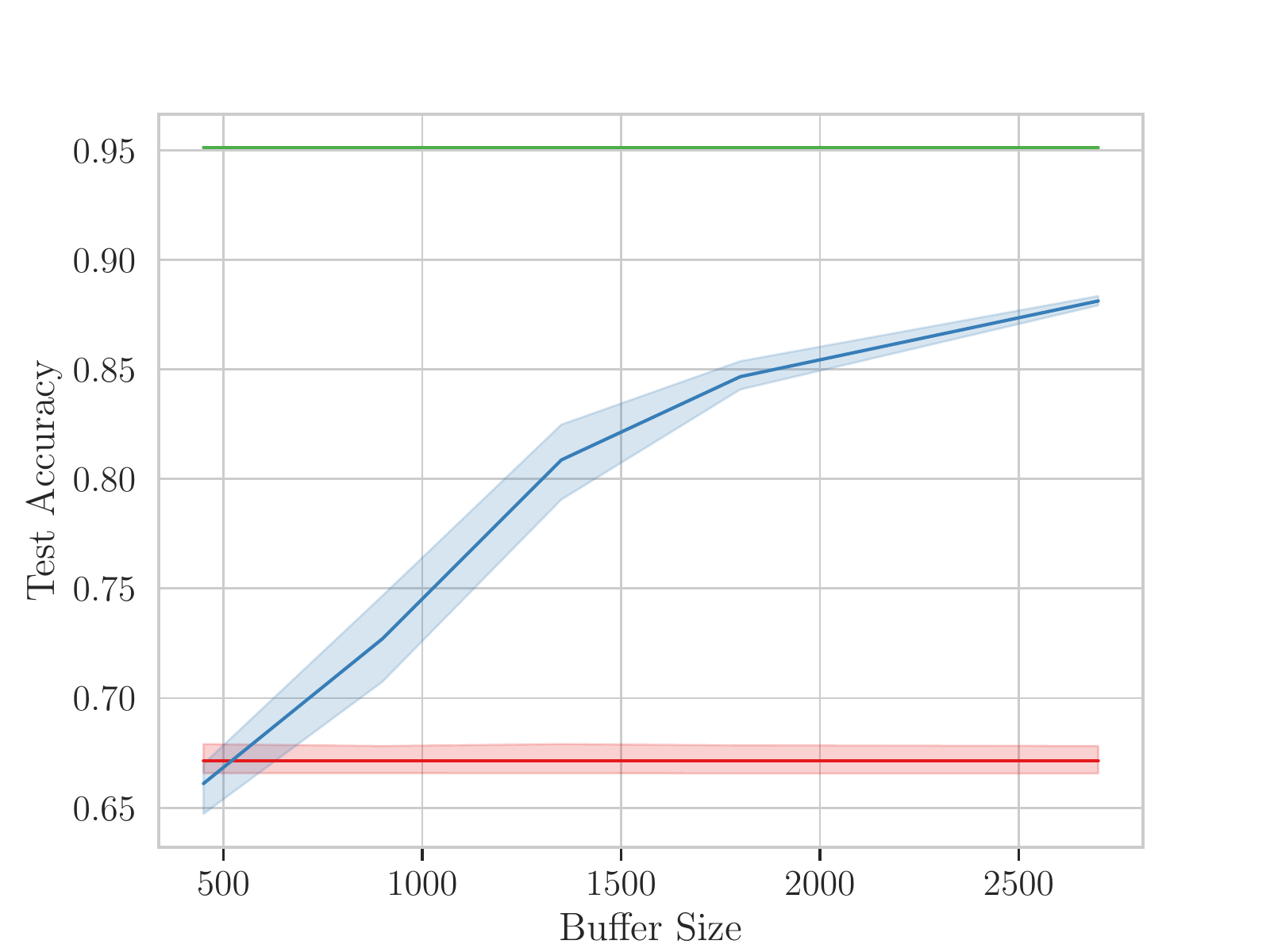}
         \caption{GelSight-45}
         \label{fig:gel45_cl_buff}
     \end{subfigure}
     \caption{CL performance across all BioTac-50, RoboSkin-20, and GelSight-45 datasets, for varying buffer sizes. Accuracy from supervised upper bound and ridge regression are shown to illustrate the performance changes associated with adaptation. With increasing memory buffer, CL achieves better test accuracy and narrows the gap against standard supervised learning.}
     \label{fig:cl_buff}
\end{figure*}

\cref{fig:random_noise_dir} shows that model trained without random jittering augmentation generalizes poorly to noisy data, especially on BioTac dataset. This is due to the BioTac data being collected under a strict condition, including fixed force and movement speed. The model trained on non-augmented BioTac data thus overfits to the homogeneous data and lack robustness. In contrast, RoboSkin data contains more diverse samples since it is collected without strict speed or force control. As reflected in \cref{fig:random_noise_dir}, the non-augmented model trained on RoboSkin data is therefore naturally robust to a low level of sensor noise. However, as the noise level increases, the test accuracy of all non-augmented models still deteriorate rapidly.

\cref{fig:random_noise_dir} also indicates that the model trained with augmentation can significant sensor noise, with the noise level of 0.5 representing a potentially 50\% deviation from the intended value range. At this level, the augmented model still retains a test accuracy of 80\% for RoboSkin and 73\% for BioTac-50. Lastly, we observe that even for the original test set (i.e, noise level = 0), the augmented model still outperforms the non-augmented version, suggesting more robust features learned with augmentation.

Overall, we have demonstrated that standard CV augmentations can be directly applied to tactile images to appreciably boost model robustness in various aspects, including sampling length,  movement speed and sensor noise. As several of our experiments relied on simulated test data, we will further demonstrate the usefulness of augmentation with real out-of-distribution data in \cref{sec:exp_contactile}.

\subsection{Continual Tactile Representation Learning}

As described in \cref{sec:bg_tasks}, we cast material classification in a CL setting, which requires our model to learn each material sequentially. CL enables robots to continuously acquire new tactile experiences, without having to perform expensive retraining from scratch. 


\paragraph{Model Detail} The same foundational vision model is used as the embedding model for \cref{alg:scroll}. During fine-tuning with memory buffer $M_t$, we adopt data augmentation and a cosine learning schedule~\cite{loshchilov2016sgdr} to mitigate overfitting. For all experiments, we perform a 5-fold cross-validation.


\cref{fig:cl_buff} shows the CL performance for each dataset over different memory buffer sizes. We report the performance of $f_t$ and the fine-tuned $f_t^*$. We also include the test accuracy of standard material classification as a performance reference. Note that $f_t$ obtained via recursive least squares is equivalent to the least-squares baseline discussed in \cref{sec:mat_class}. Thanks to the foundational vision model, $f_t$ thus guarantees a robust minimum performance level for CL (see red lines in \cref{fig:cl_buff}). $f_t^*$ is obtained by adapting $f_t$ with the memory buffer. Its performance improves with larger memory buffers, closing the gap with standard material classification. For BioTac and RoboSkin particularly, the CL performance is comparable with standard supervised learning, using a moderate memory buffer of 1500 and 600 respectively. The memory buffer required only represents a fraction of the original datasets, suggesting that our approach also allows efficient and accurate CL of new materials with limited memory requirements.

\subsection{Fabric Composition Detection}
\label{sec:exp_contactile}
Introduced in \cref{sec:bg_tasks}, fabric composition detection involves predicting the presence of six constituent materials, including Linen, Viscose, Cotton, Wool, Polyester and Elastane, in different fabrics. A single model is learned to detect the presence of all constituents concurrently, with one prediction head for each constituent. This task is more challenging than standard material classification, due to the ``similar feels'' of different fabrics. The physical weave of a fabric also contributes to its feel, adding a potential confounding factor for the task.

For this task, the data is collected using Contactile sensor. As discussed in \cref{sec:sensor_data}, we deliberately used two protocols for data collection. The training set is collected using strict force and velocity control while the test set is collected with more natural movements. The test set thus presents a more realistic setting and a clear domain shift with respect to the training data.

\paragraph{Model Details} The training procedure is similar to that used for standard material classification. The only change is that the number of training epochs is reduced from 100 to 50. Data augmentations are applied to model training when specified. For evaluation, we consider the average classification score for all constituents materials. For instance, Felt contains Viscose and Wool. The learned model only achieves a score of 1 for predicting precisely the two constituents. Any false positive or false negative detection will decrease the score by $\frac{1}{6}$.

\cref{tab:Contactile-b} shows the average classification score for different model setups. We investigate both knowledge transfer from foundational vision model and model pre-trained on other sensors. We also study the effects of data augmentation.

\begin{table}[h]
\centering
\caption{Fabric Composition Detection Accuracy (\%)}
\label{tab:Contactile-b}
\begin{tabular}{lc}
\midrule
\textbf{Model}                       & \textbf{Test Accuracy Score} \\ \midrule\midrule
Least Squares w/ vision Pre-train  & $74.2$                  \\
Least Squares w/ BioTac Pre-Train & $76.1$                  \\
\hdashline[1pt/1pt]\noalign{\vskip 1.0ex}
ResNet              & $76.3$                  \\
ResNet + Augmentation   & $78.9$                 \\ 
ResNet + Augmentation (BioTac Pre-train)   & $80.6$                 \\ \midrule
\end{tabular}
\end{table}

In \cref{tab:Contactile-b}, we again leverages least-squares classifier over a fixed representation to quantify the effectiveness of a pre-trained model. We see that directly applying the foundational vision model achieves 74.2\%, while applying the BioTac model obtained in \cref{sec:mat_class} achieves 76.1\%. The result is our first demonstration of \textit{successful cross-task and cross-sensor transfer}: the BioTac model trained on standard material classification can be directly applied to Contactile data for fabric composition detection. This result demonstrates the general applicability of our approach, and its ability for robust and flexible knowledge transfer.


\cref{tab:Contactile-b} further demonstrates the usefulness of data augmentations on real out-of-distribution data, with augmentation contributes over 2\% in test accuracy compared to the non-augmented model. The results validate our physical interpretations for the applied augmentations, showing that the augmented model is indeed more robust against more varied motions. From another perspective, we may also leverage the synthetic data produced by augmentation to reduce data collection load. This is important if a robot is only allowed limited (exploratory) interaction with environments. Lastly, we remark that the best model is obtained by combining both knowledge transfer and augmentation, achieving 80.6\% in test accuracy.

\subsection{Observations on the Learned Representation}
Results from previous sections suggest robust knowledge transfer across sensors despite the varied sensing mechanisms and data format. We hypothesize that this could be a result of a learned invariant descriptor of the tactile properties of the contact surfaces. Since the processing of texture in the human somatosensory cortex is a relatively lower-level function, we are thus interested in understanding if the lower-level abstraction in the learned model recovers similar latent representation for diverse sensor data.

\begin{figure}[h]
     \centering
     \begin{subfigure}[b]{0.46\linewidth}
         \centering
         \includegraphics[width=\linewidth]{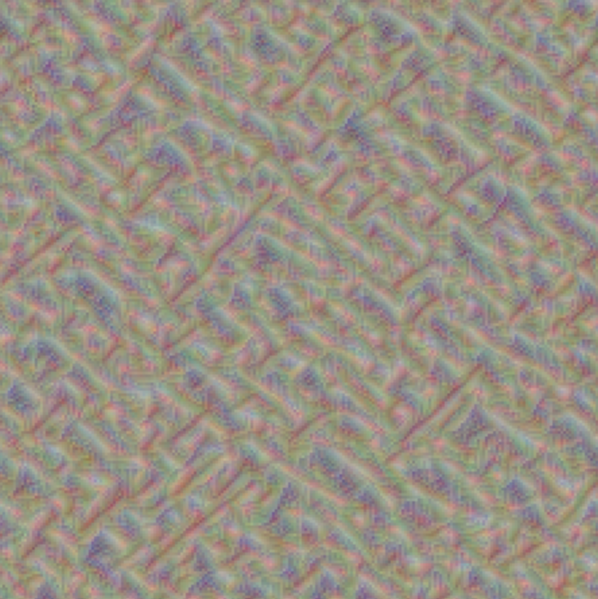}
         \caption{Foundational Model}
     \end{subfigure}
     \hfill
     \begin{subfigure}[b]{0.46\linewidth}
         \centering
         \includegraphics[width=\linewidth]{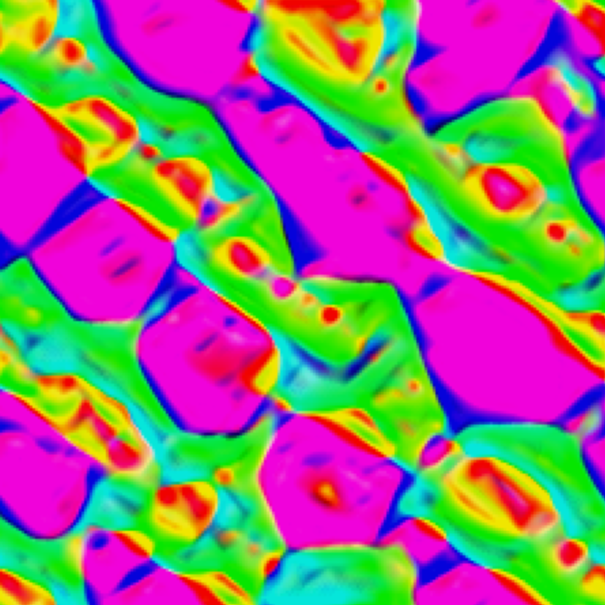}
         \caption{GelSight}
     \end{subfigure}
     \qquad
    \begin{subfigure}[b]{0.46\linewidth}
         \centering
         \includegraphics[width=\linewidth]{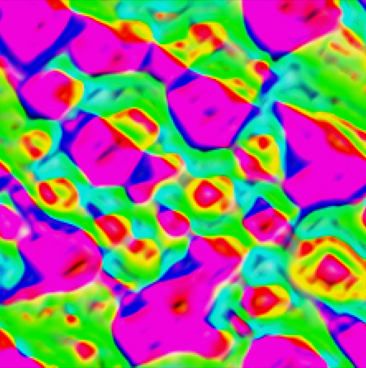}
         \caption{BioTac-50}
     \end{subfigure}
     \hfill
     \begin{subfigure}[b]{0.46\linewidth}
         \centering
         \includegraphics[width=\linewidth]{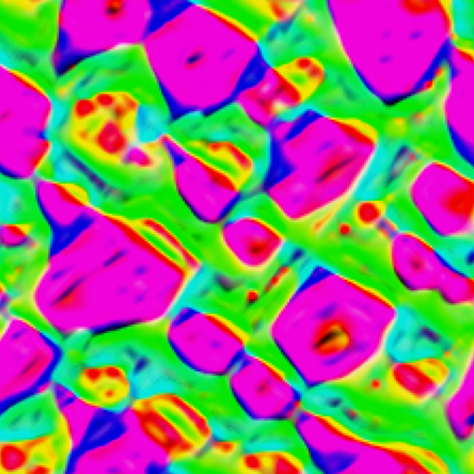}
         \caption{RoboSkin}
     \end{subfigure}
     \caption{Feature activation after block 1 of ResNet. \textbf{(a)} Original feature activation from foundational CV model. \textbf{(b)}, \textbf{(c)}, \textbf{(d)} Feature activation after fine-tuning with specific sensor data.}
     \label{fig:activation}
\end{figure}

\cref{fig:activation} shows the feature activation for different sensors using Deep Dream technique~\cite{Mordvintsev2015InceptionismGD}. This qualitative visualization of the learned features shows that feature activation generated right after the first block for 3 ResNets, each fine-tuned on a separate tactile dataset in standard material classification. All 3 feature activation maps have high resemblance of one another, suggesting that learned model indeed recovers consistent representation of tactile properties despite diverse sensing mechanisms. This further supports the knowledge transferrability between different sensors and related tasks.


\section{Conclusion} 
\label{sec:conclusion}
In this work, we presented a foundational model approach to tactile representation learning. In contrast to sensor-specific tactile models, our approach is characterized by a standardized ML pipeline, including a unifying data format for diverse tactile data, fully shared model architecture and learning techniques, all of which are key requirements for foundational models. Further, the experiment results suggest that our approach not only outperforms sensor-specific models, but crucially allows efficient knowledge transfer between models trained on different sensors and tasks, satisfying the remaining property for foundational models. In particular, we demonstrated the connection between visual and tactile perception, showing that foundational vision models trained on natural images can be a readily accessible source of knowledge for tactile representation learning. This also allows us to effectively perform, with the same unified model, downstream tasks which were previously achieved with an array of methods in the literature. 
We believe that this investigation thus contributes a robust and general approach to tactile representation learning and provides a strong baseline for future research.



\bibliographystyle{unsrtnat}
\bibliography{references}

\end{document}